\documentclass[letterpaper, 10 pt, conference]{ieeeconf}  

\IEEEoverridecommandlockouts                              
\usepackage[linesnumbered,ruled]{algorithm2e}
\usepackage{float}
\usepackage{subfigure}
\usepackage{graphics} 
\usepackage{epsfig} 
\usepackage{times} 
\usepackage{amsmath} 
\usepackage{amssymb}  
\usepackage{color}
\usepackage[table]{xcolor}
\usepackage[T1]{fontenc}
\usepackage{bm}
\usepackage{cite}
\newcommand*{\eg}{\textit{e.g.}\@\xspace}
\newcommand*{\ie}{\textit{i.e.}\@\xspace}
\usepackage{hyperref}
\usepackage{booktabs}
\hypersetup{
	colorlinks=true,
	linkcolor=blue,
	filecolor=blue,      
	urlcolor=blue,
	citecolor=cyan,
}

\title{\LARGE \bf
	Learning Consistency Pursued Correlation Filters for\\ Real-Time UAV Tracking
}

\vspace{-5pt}
\author{Changhong Fu$^{1,*}$, Xiaoxiao Yang$^{2}$, Fan Li$^{1}$, Juntao Xu$^{1}$, Changjing Liu$^{1}$, and Peng Lu$^{3}$
	\thanks{*Corresponding author}
	\thanks{
		$^{1}$Changhong Fu, Fan Li, Juntao Xu and Changjing Liu are with the School of Mechanical Engineering, Tongji University, 201804 Shanghai, China.
		{\tt\small changhongfu@tongji.edu.cn}}%
	\thanks{$^{2}$Xiaoxiao Yang is with the School of Electronics and Information Engineering, Tongji University, 201804 Shanghai, China.}
	\thanks{$^{3}$Peng Lu is with the Adaptive Robotic Controls Lab (ArcLab), Hong Kong Polytechnic University (PolyU), Hong Kong, China	
		{\tt\small peng.lu@polyu.edu.hk}}%
}
\begin{document}
	
	\maketitle
	\thispagestyle{empty}
	\pagestyle{empty}
	
	%
	
	\begin{abstract}
		Correlation filter (CF)-based methods have demonstrated exceptional performance in visual object tracking for unmanned aerial vehicle (UAV) applications, but suffer from the undesirable boundary effect.
	 To solve this issue, spatially regularized correlation filters (SRDCF) proposes the spatial regularization to penalize filter coefficients, thereby significantly improving the tracking performance.
		However, the temporal information hidden in the response maps is not considered in SRDCF, which limits the discriminative power and the robustness for accurate tracking.
		This work proposes a novel approach with dynamic consistency pursued correlation filters, \ie, the CPCF tracker. 
		Specifically, through a correlation operation between adjacent response maps, a practical consistency map is generated to represent the consistency level across frames.
		By minimizing the difference between the practical and the scheduled ideal consistency map, the consistency level is constrained to maintain temporal smoothness, and rich temporal information contained in response maps is introduced.
		Besides, a dynamic constraint strategy is proposed to further improve the adaptability of the proposed tracker in complex situations.
		Comprehensive experiments are conducted on three challenging UAV benchmarks, \ie, UAV123@10FPS, UAVDT, and DTB70. Based on the experimental results, the proposed tracker favorably surpasses the other 25 state-of-the-art trackers with real-time running speed ($\sim$43FPS) on a single CPU. 	


	\end{abstract}
	\section{INTRODUCTION}
	Nowadays, due to the unmatched mobility and portability, unmanned aerial vehicle (UAV) has aroused widespread attention for various applications, such as path planning~\cite{Laguna2019IROS}, autonomous landing~\cite{Fu2014ICUAS}, obstacle avoidance~\cite{Fu2014ICRA}, and aerial cinematography~\cite{Bonatti2019}.
	As the basis of the above applications, developing a real-time, robust and accurate tracking method is imperative. 
	However, due to many challenges introduced by unmanned airborne flight, such as aggressive UAV motion and viewpoint change, visual tracking in UAV applications is still a tough task.
	Besides, the nature of UAV also presents great challenges for visual tracking, \eg, mechanical vibration, limited computing power, and battery capacity.
	
		\begin{figure}[t]
		\centering
		\includegraphics[width=0.485\textwidth]{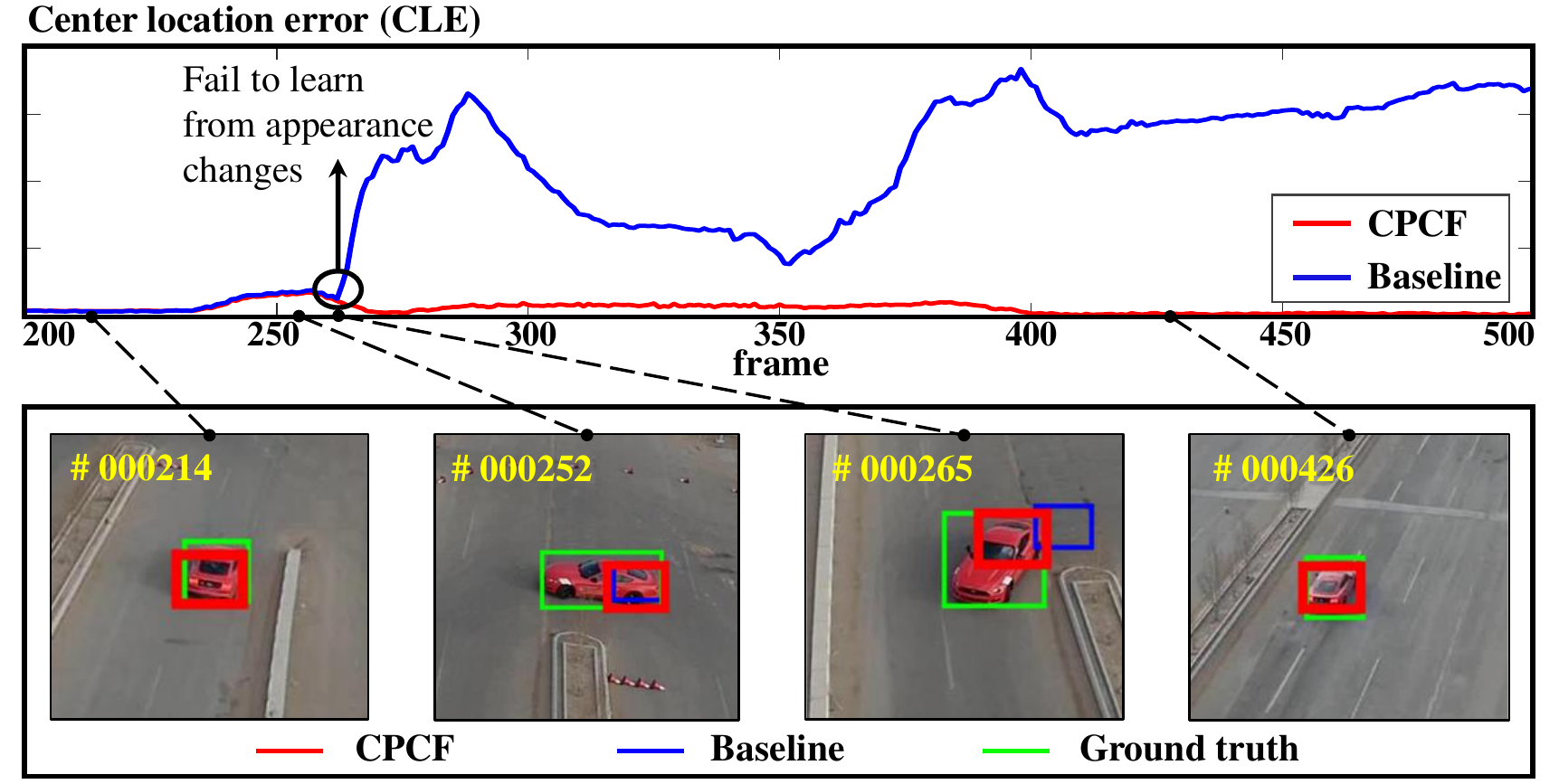}
		\vspace{-20pt}
		\caption{
		Center location error (CLE) comparison between the presented CPCF tracker and the baseline, \ie, SRDCF \cite{Danelljan2015ICCV}.  
		CLE denotes the Euclidean distance between the estimated center location of the target and the ground truth. 
		As shown in the sequence \textit{car16\_2} from UAV123@10FPS, when the viewpoint change occurs, CPCF shows great adaptability for appearance changes, while the baseline fails to track the object robustly.  }
		\label{fig1}
	\end{figure}

	Correlation filter (CF)-based approaches~\cite{Henriques2015TPAMI, Galoogahi2017CVPR, Fu2019IROS, Fu2020TGRS} have been extensively applied to tackle with the aforementioned problems in UAV tracking, due to the high computational efficiency and satisfactory tracking performance. 
	By using the property of a circular matrix, the CF-based methods can learn correlation filters efficiently in the frequency domain. 
	The number of negative samples also increases significantly without producing a heavy computational burden. 
	Nonetheless, due to the property of cyclic shift operation, inaccurate negative samples are introduced by the undesired boundary effect, which substantially reduces the discriminative power of the learned model.  
	To tackle this problem, spatially regularized correlation filters (SRDCF)~\cite{Danelljan2015ICCV} proposes the spatial regularization to penalize the filter coefficients in the background.
	In this way, the boundary effect is mitigated and a larger set of negative samples are introduced, which significantly improves the tracking performance.
	However, during a practical tracking process, the continuity between image frames implies a strong time sequence correlation worth exploring. 
	Essentially, the time sequence correlation is caused by the continuity of the object location in the image, which is finally reflected as the continuity of response maps in the time domain.
	Therefore, the introduction of response maps has been a crucial issue for exploiting temporal information efficiently.
	Notwithstanding, SRDCF focuses on improving spatial solutions without concerning the key temporal information in response maps.

	\vspace{-2pt}	
	To thoroughly explore temporal information in response maps without losing computation efficiency, this work proposes to pursue dynamic consistency across frames.
	Specifically, with the correlation operation between response maps, the consistency map is produced to evaluate the consistency level for consecutive two frames.
	Furthermore, an ideal consistency map with the highest consistency level is designed as the correlation result between two ideal responses.
	By minimizing the difference between the ideal and the practical consistency map, the consistency is forced to maintain a high level, and thus the rich temporal information is injected efficiently.
	Moreover, considering the rate of appearance changes is distinct in various tracking periods, a dynamic constraint is introduced to avoid the mismatch between the fixed ideal consistency map and the rate of appearance changes. 
	Concretely, depending on the quality of response maps, the ideal consistency map is dynamically adjusted to meet the requirement of consistency level and further enhance the adaptiveness in different UAV tracking scenarios.
	As shown in Fig.~\ref{fig1}, in case of viewpoint change, the CPCF tracker is well adapted to the fast appearance changes, while the baseline fails to track the target robustly. 
	
	Therefore, a novel approach with dynamic \textbf{c}onsistency \textbf{p}ursued \textbf{c}orrelation \textbf{f}ilters is proposed, \ie, the CPCF tracker. 
	The main contributions of this work are as follows:
	\begin{itemize}
		\item A novel method to pursue consistency across frames is proposed. 
		In this way, rich temporal information in response maps is exploited thoroughly to boost the accuracy and robustness in the UAV tracking process. 
		\item A dynamic constraint strategy is introduced to set up an adaptive restriction on the consistency level. 
		Based on the quality of the previous response map, the dynamic constraint can adaptively adjust a suitable consistency level and further increases the flexibility to cope with object appearance changes in UAV tracking. 
		\item The CPCF tracker is evaluated exhaustively on three challenging UAV benchmarks. 
		It is compared with 25 state-of-the-art trackers including both hand-crafted and deep trackers. 
		Experiments verify that the CPCF tracker favorably surpasses other trackers in terms of both accuracy and robustness with satisfactory speed for real-time tasks on a single CPU.
	\end{itemize}

	\begin{figure*}[!h]
		\centering
		\includegraphics[width=1\textwidth]{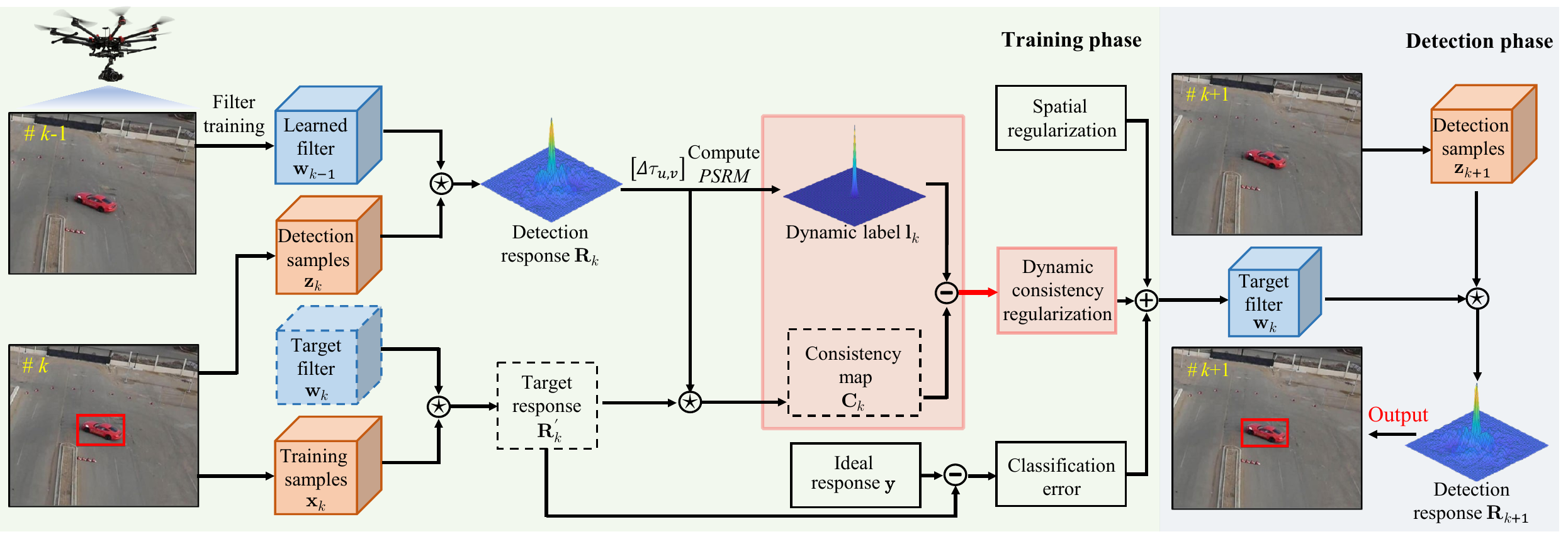}
		\vspace{-25pt}
		\caption{ The main workflow of the proposed CPCF tracker. 
		In the training phase, to evaluate the consistency level between frames, the consistency map $\mathbf{C}_k$ is generated by responses $\mathbf{R}_{k}$ and $\mathbf{R}_k^{'}$ via the correlation operation.
		Moreover, based on the $PSRM$ scores calculated from $\mathbf{R}_{k}$, the label $\mathbf{l}_k$ is dynamically adjusted to set a self-adaptive constraint on $\mathbf{C}_k$.
		Combing the spatial regularization, classification error and dynamic consistency regularization, the filter training objective is solved through an ADMM optimization to obtain filter $\mathbf{w}_k$.
		Note that in the training phase of the $k$-th frame, the to be obtained the target filter $\mathbf{w}_k$, the target response $\mathbf{R}_k$ and the consistency map $\mathbf{C}_k$ are represented by dashed lines.}
		\label{fig:mainstructure}
	\end{figure*}
	\vspace{-5pt}
	\section{RELATED WORKS}\label{sec:RELATEDWORK}
	\vspace{-2pt}
	\subsection{Tracking with correlation filters}
	\vspace{-2pt}
	CF-based approaches have been widely applied in visual object tracking tasks since the proposal of the minimum output sum of the squared error (MOSSE) filter~\cite{Bolme2010CVPR}.
	J. F. Henriques \textit{et al}.~\cite{Henriques2015TPAMI} extend MOSSE by exploiting the kernel trick and multi-channel features to improve the CF-based method. 
	Besides, the CF-based framework is further developed by multi-resolution scale~\cite{Li2014ECCVws} and part-based analysis~\cite{YLi2018CVPR,Fu2019ACCESS}.
	For feature extraction methods, the hand-crafted features including histogram of oriented gradient (HOG)~\cite{Dalal2005CVPR} and color names (CN)~\cite{Danelljan2014CVPR} have been widely used in the tracking process.
	Moreover, to attain a more comprehensive object appearance representation, some recent works~\cite{Li2020TMM,Danelljan2015ICCV} have combined deep features into the CF-based framework. 
	Nonetheless, the heavy computational load brought by deep features deprives it of the ability to be applied in real-time UAV tracking tasks. 
	Consequently, it is still an open problem to design a tracker with both outstanding performance and satisfactory running speed.
	
	\subsection{Tracking with spatial information}
	To improve both the tracking accuracy and robustness, recent methods utilizing spatial information have been proposed~\cite{Danelljan2015ICCV,Danelljan2017CVPR,Danelljan2016ECCV,Galoogahi2017CVPR}.
	By integrating the spatial regularization, SRDCF~\cite{Danelljan2015ICCV} can penalize the background representing filter coefficients and learn the filter on a significantly larger set of negative training samples.
	Background-aware correlation filter (BACF)~\cite{Galoogahi2017CVPR} directly multiplies the filter with a binary matrix to expend the search regions. 
	In this way, BACF can utilize not only the target but also the real background information for training.
	In the CSR-DCF tracker~\cite{Lukezic2017CVPR}, the filter is equipped with spatial reliability maps to improve the tracking of non-rectangular targets and suppresses the boundary effects.
	However, the improvement brought by spatial information alone is not enough comprehensive. 
	In addition to the spatial information, the effective introduction of both spatial and temporal information has attracted increasing attention among the CF-based tracking community. 
	
	\vspace{-3pt}
	\subsection{Tracking with temporal information}
	Considering the strong time sequence correlation between the video frames, some trackers exploit the temporal information to further improve the tracking performance~\cite{Danelljan2016CVPR,Guo2017ICCV, Li2018CVPR}. 
    SRDCFdecon~\cite{Danelljan2016CVPR} reweights its historical training samples to reduce the problem caused by sample corruption. 
	However, depending on the size of the training set, the tracker may need to store and process a great number of historical samples and thereby sacrificing its tracking efficiency.
	STRCF~\cite{Li2018CVPR} proposes a temporal regularization to penalize the variation of filter coefficients in an element-wise manner and ensures the temporal smoothness. 
	However, the rigid element-wise constraint may fail the filter in learning critical appearance changes and limit the adaptiveness of the tracker.
	Thus, the presented CPCF tracker considers the temporal information by evaluating the consistency level between response maps as a whole. 
	Consequently, the CPCF tracker can maintain the temporal smoothness flexibly and enhance the robustness of tracking.


	\section{PROPOSED TRACKING APPROACH}\label{sec:METHOD}
	\vspace{-1pt}
	In this work, the proposed CPCF tracker focuses on pursuing the dynamic consistency across frames. 
	Therefore, this section first introduces the consistency evaluation method, then introduces the design process of the dynamic consistency constraint. 
	Finally, the overall objective of CPCF is given. 
	Its main workflow can be seen in Fig.~\ref{fig:mainstructure}.
	
	\vspace{-3pt}
	\subsection{Consistency evaluation}
	\label{sec:Consistency Evaluation}
	To evaluate the consistency across frames, this work proposes to study the similarity between the detection response $\mathbf{R}_k$ and target response $\mathbf{R}_k^{'}$.
	Specifically, the responses $\mathbf{R}_k$ and $\mathbf{R}_k^{'}$ are obtained as:
	\begin{equation}
	\begin{cases}
	\mathbf{R}_k =  \sum _ { d = 1 } ^ { D } \mathbf { w }_{k-1} ^ { d } \star \mathbf { z }_k ^ { d }
	\\
	\mathbf{R}_k^{'} = \sum _ { d = 1 } ^ { D } \mathbf { w }_k ^ { d } \star \mathbf { x }_k ^ { d }
	\end{cases}
	\ ,
	\end{equation}
	
	\noindent where the subscripts $(k-1)$ and $k$ denote the $(k-1)$-th and $k$-th frame, respectively. 
	The superscript $d$ denotes the $d$-th channel.
	$\mathbf{w}_{k-1}^d$ and $\mathbf{w}_k^d$ denote the correlation filters.
	$\mathbf{z}_{k}^d$ and $\mathbf{x}_k^d$ denote the detection samples and training samples. 
	The operator $\star$ denotes the cyclic correlation operation.

	Considering the potent time sequence correlation between the detection response $\mathbf{R}_k$ and the target response $\mathbf{R}_k^{'}$, bountiful temporal information is hidden in the consistency between frames.
	Therefore, to exploit the temporal information, the cyclic correlation operation is adopted to evaluate the consistency between responses by the following formula:
	\begin{equation}\label{consistency map}
	\begin{aligned}
	\mathbf { C }_k  = \mathbf{ R }_{ k }^{'}  \star \mathbf { R }_{k}[\varDelta\tau_{u, v}]   \ ,
	\end{aligned}
	\end{equation}
	
	\vspace{-3pt}
	\noindent where $\mathbf{C}_k \in \mathbb{R}^{N}$ denotes the consistency map. 
	The subscripts $u$ and $v$ indicate the difference between the peak positions of detection response $\mathbf{R}_{k}$ and the center of the image patch. 
	The operator $[\varDelta\tau_{u, v}]$ shifts the peak of $\mathbf{R}_k$ to the center of the response map in the two-dimensional space.

	\vspace{-2pt}
	\subsection{Constraint on consistency}
	\label{sec:Constraint on Consistency}
	\vspace{-2pt}
	In traditional CF-based methods, the responses $\mathbf{R}_{k}^{'}$ and $\mathbf{R}_k$ are both forced to be equal to the ideal response $\mathbf{y}$ in the ideal situation.
	Accordingly, considering the generating process of $\mathbf{C}_k$ in Eq. (\ref{consistency map}), an ideal consistency map should be produced by two ideal responses $\mathbf{y}$. 
	Thus, the ideal consistency map can be served as the consistency constraint label for the practical consistency map $\mathbf{C}_k$, and a fixed constraint label $\mathbf{l}_f$ is designed as follows:
	\vspace{-3pt}
	\begin{equation}\label{intial_z}    
	\displaystyle\mathbf{l}_f = \mathbf{ y } \star \mathbf{y} \ ,
	\end{equation}    
	\noindent where $\mathbf{y}$ denotes the ideal response.
	Besides, since the label $\mathbf{l}_f$ is produced by two identical ideal responses $\mathbf{y}$, which means the highest strength for consistency constraint is applied to $\mathbf{C}_k$ as:
	\begin{equation}   \label{constraint}
	\Vert \mathbf { l }_f -\mathbf{C}_k  \Vert _2^2 \ .
	\end{equation}   
	
	\vspace{-3pt}
	By minimizing the squared error between $\mathbf{l}_f$ and $\mathbf{C}_k$ in Eq. (\ref{constraint}), the responses $\mathbf{R}_k$ and $\mathbf{R}_k^{'}$ are forced to pursue a high consistency level, and thus abundant temporal information is injected efficiently.

	
	
	Moreover, for the practical tracking process, since the rate of the object appearance changing varies in different tracking scenarios, the label $\mathbf{l}_f$ should adapt to the rate of appearance changes instead of enforcing a fixed consistency constraint. 
	Thus, based on the fixed constraint label in Eq.~(\ref{intial_z}), the dynamic constraint label is further proposed as follows:
	\vspace{-3pt}
	\begin{equation}    
	\displaystyle\mathbf{l}_k ={h}_{k} \mathbf{l}_f \ ,
	\end{equation}    
	\noindent where the ${h}_k$ indicates the dynamic regulatory factor to adjust the constraint strength for consistency. The design of ${h}_k$ is described in detail in Section~\ref{subsec:label_z}. 
	
	\vspace{-4pt}
	\subsection{Overall objective}
	The overall objective of the presented CPCF tracker is to minimize the following loss function:
	\vspace{-3pt}
	\begin{equation}\label{all_objective}
	\begin{split}
	\mathcal { E } ( \mathbf { w }_k ) &= \frac { 1 } { 2 } \Vert \mathbf { y } - \sum _ { d = 1 } ^ { D } \mathbf { w }_k ^ { d } \star \mathbf { x }_k ^ { d } \Vert _2^2 +\frac{1}{2} \sum_{ d = 1 }^{D}\vert\vert\mathbf{s}_k^d\mathbf{w}_k^d\vert\vert_2^2\\
	&+\frac { \gamma } { 2 } \sum _ { d = 1 } ^ { D } \Vert \mathbf { l }_k -( \sum_{ d = 1 } ^ { D }  \mathbf { w }_k ^ { d } \star \mathbf{ x }_k ^ { d }) \star \mathbf { R }_{k}[\varDelta\tau_{u, v}]   \Vert _2^2 
	\end{split}
	\ .
	\end{equation}
	
\vspace{-2pt}
	The presented loss function contains three terms, \ie, the first classification error term, the second spatial regularization term and the last dynamic consistency regularization term. 
	For the first term, $\mathbf{x}_k^d$ and $\mathbf{w}_k^d$ denote the $d$-th channel of training sample and correlation filter, respectively. 
	For the second term, the spatial weight function $\mathbf{s}_k^d$ is introduced to mitigate the boundary effect. 
	For the third term, $\gamma$ denotes the consistency penalty. The detection response $\mathbf{R}_k$ can be expressed as $\mathbf{R}_k = \sum_{ d = 1 } ^ { D }  \mathbf { w }_{k-1} ^ { d } \star \mathbf{ z }_k ^ { d }$, and can be considered as a constant in the training process.
	\subsection{Optimization operations}\label{Optimization}
	The following optimization is derived in one-dimensional case, and can be easily extended to two-dimensional case.
	To convert Eq.~(\ref{all_objective}) to frequency domain conveniently, the equation is firstly expressed as matrix form as follows: 
	\vspace{-3pt}
	\begin{equation}\label{matrixform}
	\begin{split}
	{\mathcal { E }} ( \mathbf { w }_k)& = \frac { 1 } { 2 } \Vert {\mathbf { y }} - {\mathbf {X}}_k { \mathbf { w } }_k\vert\vert_2^2 + \frac{1}{2} \vert\vert\mathbf{s}_k\mathbf{w}_k\vert\vert_2^2\\
	&+\frac{\gamma}{2}\vert\vert  {\mathbf{l}}_k- \mathbf{R}^\top  {\mathbf{X}}_k {\mathbf{w}}_k\vert\vert_2^2  \quad\quad\quad\quad\quad\quad\ \\
	\end{split}
	\ ,
	\end{equation}
	
	\noindent where $\mathbf{y} = [\mathbf{y}(1)\cdots\mathbf{y}(N)]^\top$, $\mathbf{w}_k = [{\mathbf{w}_{k}^1}^\top\cdots{\mathbf{w}_{k}^D}^\top]^\top, \mathbf{s}_k = [{\mathbf{s}_{k}^1}^\top\cdots{\mathbf{s}_{k}^D}^\top]^\top$, and $\mathbf{X}_k=[\mathbf{x}_k[\Delta\tau_1]^\top\cdots\mathbf{x}_k[\Delta\tau_N]^\top]^\top$. 
    $\mathbf{x}_k[\Delta\tau_n]$ is the circularly shifted sample. 
	$\mathbf{R}$ denotes the circular matrix generated by the shifted detection response $\mathbf { R }_{k}[\varDelta\tau_{u, v}]$. 
	The operator ${\top}$ denotes the conjugate transpose operation. 
	In order to improve computing efficiency, Eq.~(\ref{matrixform}) is further transferred into the frequency domain as follows:
	\vspace{-3pt}
	\begin{equation}\label{frequency domain}
	\begin{split}
	\hat{\mathcal { E }} ( \mathbf { w }_k, \hat{\mathbf{g}}_k)& = \frac { 1 } { 2 } \Vert \hat{\mathbf { y }} - \hat{\mathbf {X}}_k \hat{ \mathbf { g } }_k\vert\vert_2^2 + \frac{1}{2} \vert\vert\mathbf{s}_k\mathbf{w}_k\vert\vert_2^2\\
	&+\frac{\gamma}{2}\vert\vert \hat {\mathbf{l}}_k- \hat{\mathbf{X}}_k^r \hat{\mathbf{g}}_k\vert\vert_2^2  \quad\quad\quad\quad\quad\quad\ ,\\
	&s.t. \quad \hat{\mathbf{g}}_k = \sqrt{N}\mathbf{F}\mathbf{w}_k 
	\end{split}
	\end{equation}
	\noindent where $\hat{\mathbf{g}}_k \in \mathbb{C}^{DN\times1}$ is introduced as an auxiliary variable.
	The superscript\ $\hat{}$\ denotes the discrete Fourier Transform (DFT) of a signal, \ie, $\hat{\mathbf{\alpha}} = \sqrt{N}F\mathbf{\alpha}$.
	The matrix $\hat{\mathbf {X}}_k$ and  $\hat{\mathbf {X}}_k^r$ are defined as $\hat{\mathbf {X}}_k = [diag(\hat{\mathbf{x}}_k^1)^\top, \cdots, diag(\hat{\mathbf{x}}_k^D)^\top]$ of size $N\times DN$,  $\hat{\mathbf {X}}_k^r = [diag(\hat{\mathbf{r}}_{k}^1\odot\hat{\mathbf{x}}_k^1)^\top, \cdots, diag(\hat{\mathbf{r}}_{k}^1\odot\hat{\mathbf{x}}_k^D)^\top]$ of size $N\times DN$, respectively. 
	The operator $\odot$ indicates the element-wise multiplication. 
	$\hat{\mathbf{r}}_{k}$ is the discrete Fourier Transform of shifted detection response $\mathbf{R}_{k}[\varDelta\tau_{u,v}]$.

	Considering the convexity of Eq. (\ref{frequency domain}), alternative direction method of multipliers (ADMM) is introduced to achieve a globally optimal solution efficiently. Hence Eq. (\ref{frequency domain}) can be expressed in augmented Lagrangian form as follows:
	\begin{equation}\label{equ:L(w,g,zeta)}
	\begin{split}
	\mathcal{L}(\mathbf{w}, \hat{\mathbf{g}}, \hat{\zeta})
	& = \frac{1}{2} \Vert  \hat{\mathbf { y }} - \hat{\mathbf {X}}_k \hat{ \mathbf { g } }_k\Vert _2^2 + \frac{1}{2} \vert\vert\mathbf{s}_k\mathbf{w}_k\vert\vert_2^2 \\
	&+ \frac{\gamma}{2}\vert\vert \hat {\mathbf{l}}_k- \hat{\mathbf{X}}_k^r \hat{\mathbf{g}}_k\vert\vert_2^2 \\
	&+ \hat{\zeta}^\top(\hat{\mathbf{g}}_k - \sqrt{N}\mathbf{F}\mathbf{w}_k)\\
	&+ \frac{\nu}{2} \vert\vert \hat{\mathbf{g}}_k-\sqrt{N}\mathbf{F}\mathbf{w}_k\vert\vert_2^2 
	\end{split}
	\ ,
	\end{equation}
	\noindent where $\hat{\zeta} \in \mathbb{C}^{DN\times 1}$ denotes the Lagrangian vector in the Fourier domain which is defined as $\hat{\zeta} = [\hat{\zeta}^{1\top}, \cdots, \hat{\zeta}^{D\top}]^\top$ and $\nu$ denotes a penalty factor. 
	To learn filters for the ($k+1$)-th frame, ADMM algorithm should be adopted in the $k$-th frame. 
	The augmented Lagrangian form can be solved by alternatingly solving subproblems $\hat{\mathbf{g}}_{k+1}^*$ and $\mathbf{w}_{k+1}^*$ as follows:
	
	\subsubsection{Subproblem $\mathbf{w}_{k+1}^*$}
	\begin{equation}\label{subproblem_w}
	\begin{split}
	\small
	\mathbf { w }_{k+1} ^ { * } = &   \mathop{\arg \min} _ { \mathbf{ w }_k } \left\{  \right.   
	\frac{1}{2} \vert\vert\mathbf{s}_k\mathbf{w}_k\vert\vert_2^2 +\hat{\zeta}^\top \big( \hat{\mathbf{g}}_k - \sqrt{N}\mathbf{F}\mathbf{w}_k \big)\\
	& \quad + \frac{\nu}{2} \Vert \hat{\mathbf{g}}_k - \sqrt{N}\mathbf{F}\mathbf{w}_k\Vert_2^2 \} \\ 
	& \quad = (\mathbf{S}^\top\mathbf{S}+\nu)^{-1}(\zeta+\nu \mathbf{g})
	\end{split}
	\ ,
	\end{equation}
	\noindent where $\mathbf{S}$ denotes the $DMN\times DMN$ diagonal matrix concatenating $D$ diagonal matrices $diag(\mathbf{s})$.	
	
	\subsubsection{Subproblem $\hat{\mathbf{g}}_{k+1}^*$}
	\begin{equation}\label{subproblem_g}
	\begin{split} 
	\hat{\mathbf{g}}_{k+1}^* = & \mathop{\arg\min}\limits_{\hat{\mathbf{g}}_{k}}
	\{\frac { 1 } { 2 } \Vert \hat{\mathbf { y }} - \hat{\mathbf {X}}_k \hat{ \mathbf { g } }_k\vert\vert_2^2 \\
	& \quad +\frac{\gamma}{2}\vert\vert \hat {\mathbf{l}}_k- \hat{\mathbf{X}}_k^r \hat{\mathbf{g}}_k\vert\vert_2^2\\
	& \quad + \hat{\zeta}^\top \big( \hat{\mathbf{g}}_k - \sqrt{N}\mathbf{F}\mathbf{w}_k \big)\\
	& \quad + \frac{\nu}{2} \Vert \hat{\mathbf{g}}_k - \sqrt{N}\mathbf{F}\mathbf{w}_k\Vert_2^2 \}
	\end{split}
	\ ,
	\end{equation}
	
	\noindent solving the subproblem $\mathbf{g}_{k+1}^*$ directly can bring heavy computational burden due to $\hat{\mathbf {X}}_k^r \hat{ \mathbf { g } }_{k}$ and $\hat{\mathbf {X}}_k \hat{ \mathbf { g } }_k$ in the function. 
	Fortunately, $\hat{\mathbf {X}}_k$ and  $\hat{\mathbf {X}}_k^r$ are sparse banded, and thus each element in $\hat{\mathbf{y}}$, \ie, $\hat {\mathbf{y}}(n) (n=1,2,...,N)$ is only dependent on each $\hat{\mathbf{x}}_k(n)=\left[\hat {\mathbf{x}}_k^1(n),\hat {\mathbf{x}}_k^2(n),...,\hat {\mathbf{x}}_k^D(n)\right]^{\top}$ and $\hat { \mathbf { g } }_k ( n ) = \left[  { conj } \left( \hat { \mathbf { g } }_k ^ { 1 } ( n ) \right) , \ldots , { conj } \left( \hat { \mathbf { g } }_k ^ { D } ( n ) \right) \right] ^ { \top }$. 
	The operator $conj(.)$ denotes the complex conjugate.
	Therefore, the subproblem $\mathbf{g}_{k+1}^*$ can be divided into $N$ independent objectives as:
	\vspace{-5pt}
	\begin{equation}\label{gkn}
	\begin{split} 
	\small
	&\hat { \mathbf { g } }_{k+1}^* ( n ) ^ { } =  \mathop{\arg \min} _ { \hat { \mathbf { g } }_{k} ( n ) } \{ \frac { 1 } { 2 } \Vert \hat { \mathbf { y } } ( n ) - \hat { \mathbf { x }}_k  ( n ) ^ { \top } \hat { \mathbf { g } }_k ( n ) \Vert_2^2  \\ 
	&\quad\quad\quad\quad\quad\quad\quad+ \frac { \gamma } { 2 } \Vert \hat { \mathbf { l } }_k ( n ) - \hat { \mathbf { x }}_k^r ( n ) ^ { \top } \hat { \mathbf { g } }_k ( n ) \Vert_2^2  \\ 
	&\quad\quad\quad\quad\quad \quad\quad+\hat { \zeta } ( n ) ^ { \top } ( \hat { \mathbf { g } }_k ( n ) - \hat { \mathbf { w } }_k ( n ) ) \\ 
	&\quad\quad\quad\quad\quad\quad\quad + \frac { \nu } { 2 } \| \hat { \mathbf { g } }_k ( n ) - \hat { \mathbf { w } }_k ( n ) \| _ { 2 } ^ { 2 } \} 
	\end{split}
	\ ,
	\end{equation}
	\noindent where $\hat { \mathbf { w } }_k ( n ) = \left[ \hat { \mathbf { w } }_k ^ { 1 } ( n ) , \ldots , \hat { \mathbf { w } }_k ^ { D } ( n ) \right]$ and $\hat { \mathbf { w } }_k ^ { d } = \sqrt { D } \mathbf { F } \mathbf { w } ^ { d }_k$.

	The solution to each problem is given as follows:
	\begin{equation}\label{equ:subprobg}
	\small
	\begin{split} &\hat { \mathbf { g } } _{k+1}( n ) ^ { * } \\
	&= \frac{1}{1+\gamma \hat{\mathbf{r}}(n) \hat{\mathbf{r}}(n)^\top} \left( \hat { \mathbf { x }}_k  ( n ) \hat { \mathbf { x }} _k ( n ) ^ { \top }  + \frac{\nu}{1+\gamma \hat{\mathbf{r}}(n) \hat{\mathbf{r}}(n)^\top} \mathbf { I } _ { D } \right) ^ { - 1 } \\
	&\quad \left( \hat { \mathbf { x } }_k ( n ) \hat { \mathbf { y } }_k ( n )+\gamma\hat { \mathbf { x } }_k^r ( n )\hat { \mathbf { l } }_k ( n ) -  \hat { \zeta } ( n ) + \nu \hat { \mathbf { w } } ( n ) \right) 
	\end{split}
	\ .
	\end{equation} 
	
	\noindent \textbf{Remark 1:} The solving process of Eq.  (\ref{gkn}) is shown in the appendix. 
	
	To further increase the computation efficiency, the Sherman-Morrison formula is employed, \ie, $(\mathbf{A} + \mathbf{p} \mathbf{q}^{\top}) ^ {-1} = \mathbf{A} ^ {-1} - \mathbf{A}^{-1} \mathbf{p} (\mathbf{I}_D + \mathbf{q}^{\top} \mathbf{A}^{-1} \mathbf{p}) ^ {-1} \mathbf{q}^{\top} \mathbf{A}^{-1} $. 
	Thus, Eq.~(\ref{equ:subprobg}) is identically expressed as below: 
	\vspace{-4pt}
	\begin{equation}\label{subproblem_g_result}
	\begin{aligned} 
	\small
	&\hat { \mathbf { g } }_{k+1} ( n ) ^ { * } \\
	&=  \frac{1}{\nu}\left(  \hat { \mathbf { x } }_k ( n ) \hat { \mathbf { y } } ( n )+\hat{\mathbf { x } }_k^r ( n ) \hat { \mathbf { l } }_k ( n ) - \hat { \zeta } ( n ) + \nu \hat { \mathbf { w } }_k ( n ) \right) \\
	& - \frac { \hat { \mathbf { x } }_k ( n ) } {\nu b } \left(\hat { s } _ { \mathbf { x }_k } (n)  \hat { \mathbf { y } } ( n )  -\gamma \hat { s } _ { \mathbf { x }_k } (n) \hat{\mathbf{l}}_k(n) \hat { s } _ { \zeta } ( n ) + \nu \hat { s } _ { \mathbf { w }_k } ( n ) \right)
	\end{aligned}
	\ ,
	\end{equation}
	\vspace{-4pt}
	\noindent where
	\vspace{-3pt}
	\begin{equation}
	\begin{aligned}
	\begin{cases}
	\small
	&\hat{s}_{\mathbf{x}_k}(n) = \hat { \mathbf { x } }_k ( n ) ^{\top} \hat { \mathbf { x } }_k ( n ) \\
	&\hat { s } _ { \zeta } ( n ) = \hat { \mathbf { x } }_k ( n ) ^{\top} \mathbf{\hat{\zeta}}\\ 
	&\hat { s } _ { \mathbf { w } _k} ( n ) = \hat { \mathbf { x } }_k ( n ) ^{\top} {\hat{\mathbf{ w }}_k} \\ 
	&b = \hat { \mathbf { x } }_k ( n ) ^{\top} \hat { \mathbf { x } }_k ( n ) + \displaystyle\frac{\nu}{1+\gamma \hat{\mathbf{r}}(n) \hat{\mathbf{r}}(n)^\top}\\ 
	\end{cases}
	\end{aligned}
	\ .
	\end{equation}
	
	Lagrangian parameter $\zeta$ is updated in each iteration according to the following equation:
	\vspace{-3pt}
	\begin{equation}\label{Lagragian_format}
	\mathbf{\hat{\zeta}} _ {k+1}^{j+1} = \mathbf{\hat{\zeta}} _{k+1}^ {j} + \nu \left(\mathbf{\hat{g}} ^ {*(j+1)} _{k+1} - \hat{\mathbf{w}} ^ {*(j+1)} _{k+1} \right) \ , 
	\end{equation}
	
	\noindent where the subscripts $j$ and $j+1$ indicate the $j$-th and $(j+1)$-th iteration, respectively. 
	$\mathbf{\hat{g}} ^ {*(j+1)} _{k+1}$ and $\mathbf{\hat{w}} ^ {*(j+1)} _{k+1}$ denote the solution to subproblem $\mathbf{\hat{g}} ^ {*} _{k+1}$ and $\mathbf{\hat{w}} ^ {*} _{k+1}$ in the $(j+1)$-th iteration, respectively. 
	
		\begin{figure}[t]
		\includegraphics[width=0.48\textwidth]{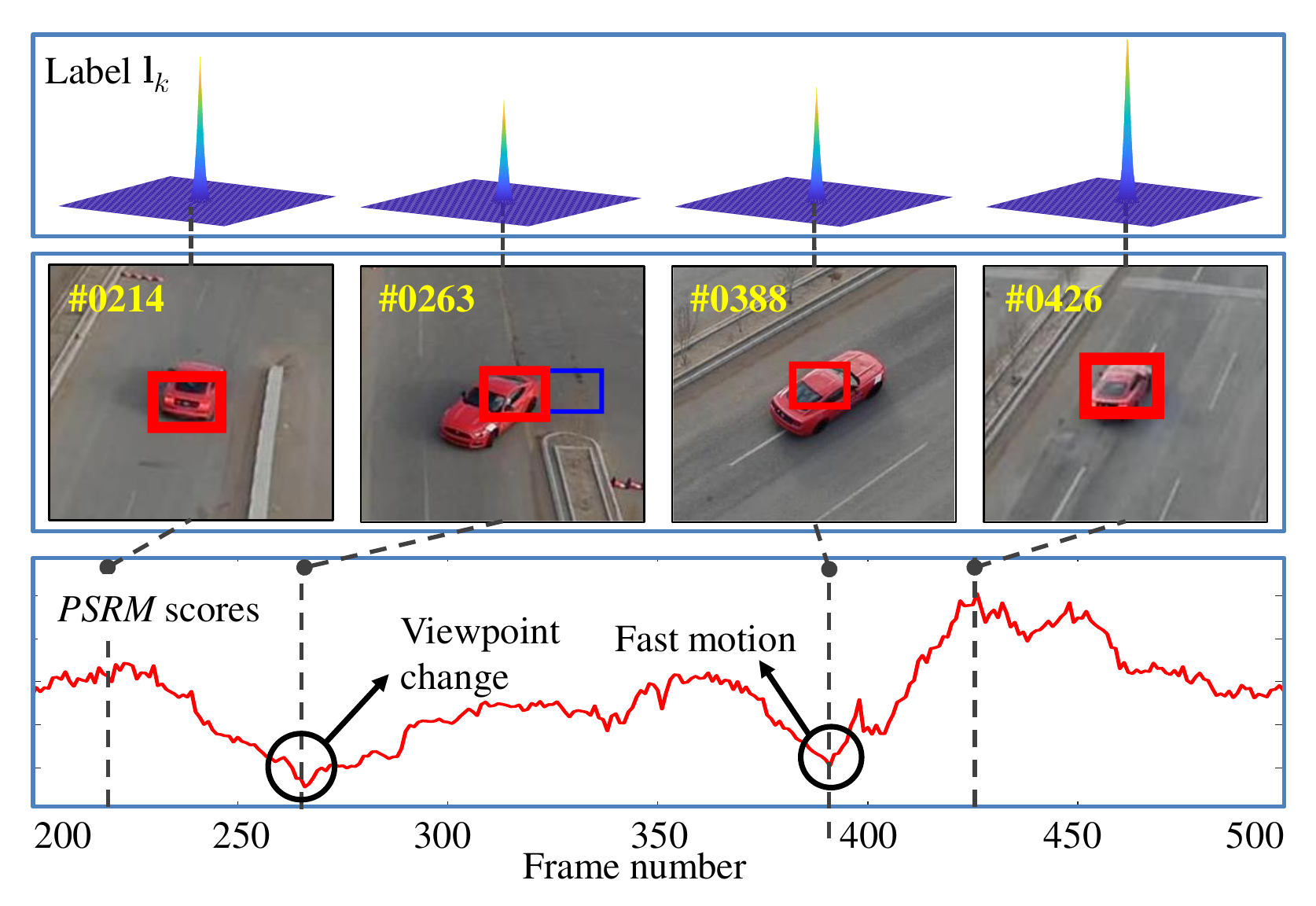}
		\vspace{-15pt}
		\caption{Illustration of the adjusting process of the dynamic constraint label $\mathbf{l}_k$. 
			The sequence \textit{car16\_2} in UAV123@10FPS benchmark is shown, and the \textcolor{blue}{blue} bounding box denotes the baseline, which fails to track the car robustly due to the fast appearance changes of the car. 
			The CPCF tracker, denoted by the \textcolor{red}{red} bounding box, appropriately adjusts the constraint strength for consistency in case of fast appearance changes and still track the target successfully.  		
		}
		\label{fig:scene}
	\end{figure}

	\subsection{Dynamic adjusting strategy for label $\mathbf{l}_k$}\label{subsec:label_z}
	In the practical tracking process, the rate of target appearance variation is various in different scenarios. 
	Therefore, an intelligent tracker should adjust the strength of consistency constraint according to the different tracking scenarios. 
	On the one hand, in terms of the fast appearance changes, constraints should be relaxed to render the tracker more authority to change and thus learn from new appearance changes. 
	On the other hand, if the appearance change is smooth, a high-level constraint for consistency is required to enhance the robustness and accuracy. 
	Thus, depending on the quality of the detection response $\mathbf{R}_k$, the dynamic regulatory factor $h_k$ is introduced as: 


	\begin{equation}\label{Eq:mag&mcer}
	{{h}}_k = {h}_{min} +\frac{{{PSRM}}}{\alpha} ({h}_{max}-{h}_{min}) \ ,
	\end{equation}
	where ${h}_{min}$ and ${h}_{max}$ denote the minimum and maximum magnitude of $\mathbf{l}_k$ respectively. $\alpha$ is a normalized coefficient. $PSRM$ denotes the quality scores of the response map as follows:
	\begin{equation}\label{PSRM}
	{{PSRM}} = \frac{R_{max} - \mu_{s1 } }{\sigma_{s1}} + \beta R_{max}\ ,
	\end{equation}
	where the first and the second term denote the peak to sidelobe ratio (PSR) and the peak value ($R_{max}$) in the response map respectively.  $\mu_{s1}$ and $\sigma_{s1}$ denote the mean value and the standard deviation in the sidelobe, respectively. $\beta$ is the weight coefficient to balance two evaluation criterions. The process of dynamic adjusting is shown in Fig. \ref{fig:scene}.

	\subsection{Model update}\label{Model Update}
	In order to improve the robustness for fast motion, viewpoint change and other challenges, an online adaptation strategy is introduced as follows:\\
	\vspace{-6pt}
	\begin{equation}\label{modelupdate}
	\hat{\mathbf{x}} _ {k}^{model} = (1-\eta) \hat{\mathbf{x}} _ {k-1}^{model}+\eta\hat{\mathbf{x}} _ {k} \ , 
	\end{equation}
	\vspace{-1pt}
	where $\eta$ is the learning rate for the appearance model. 
	$k$ and $(k-1)$ denote the $k$-th and the $(k-1)$-th frame, respectively.

\begin{figure}[!b]
	\centering
	\includegraphics[width=0.48\textwidth]{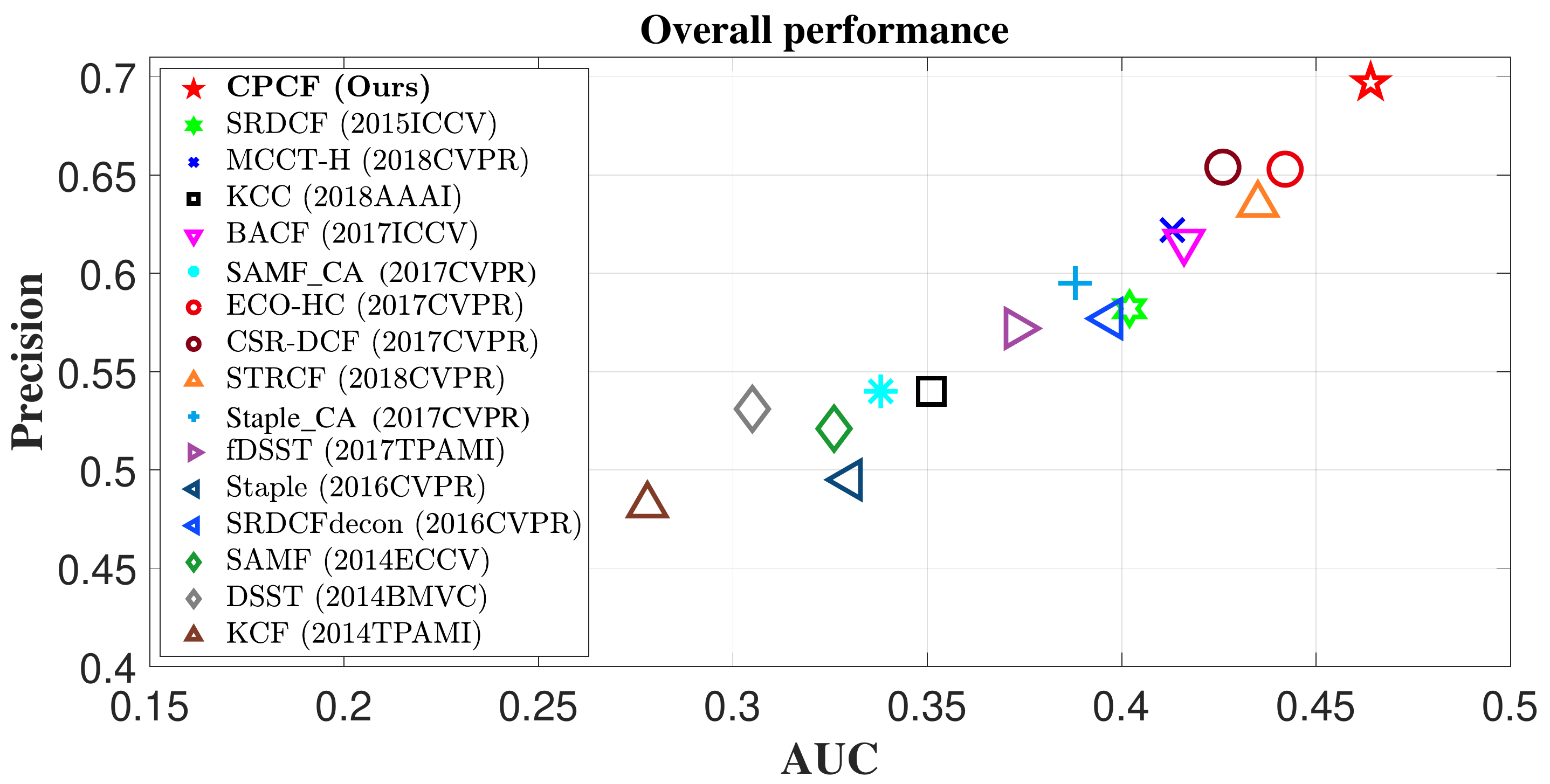}
	\vspace{-17pt}
	\caption{Comparison between the proposed CPCF tracker and other 15 state-of-the-art trackers based on hand-crafted features. The average value of precision and success rate are calculated by averaging OPE~\cite{Wu2013CVPR} result from three benchmarks.
	The CPCF tracker favorably surpasses other 15 trackers in both precision and AUC.  }
	\label{fig:stplot}
\end{figure}

\begin{figure*}[!t]\label{overall performance}
	\centering
	\setlength{\abovecaptionskip}{-2pt}
	\subfigcapskip=5pt
	\subfigure[UAV123@10FPS benchmark]{
		\begin{minipage}[t]{0.32\textwidth}
			\includegraphics[width=1\textwidth]{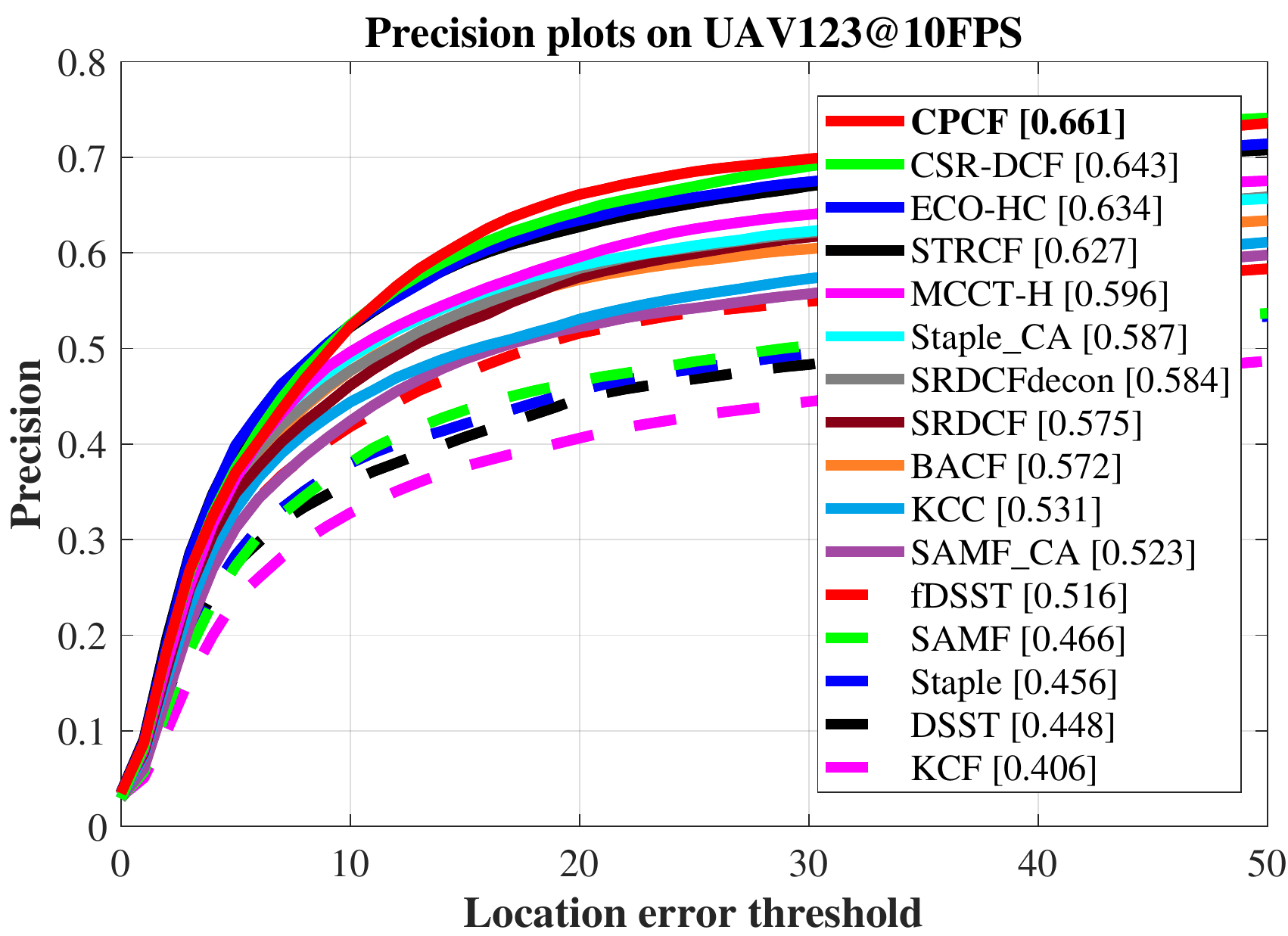}
			\includegraphics[width=1\textwidth]{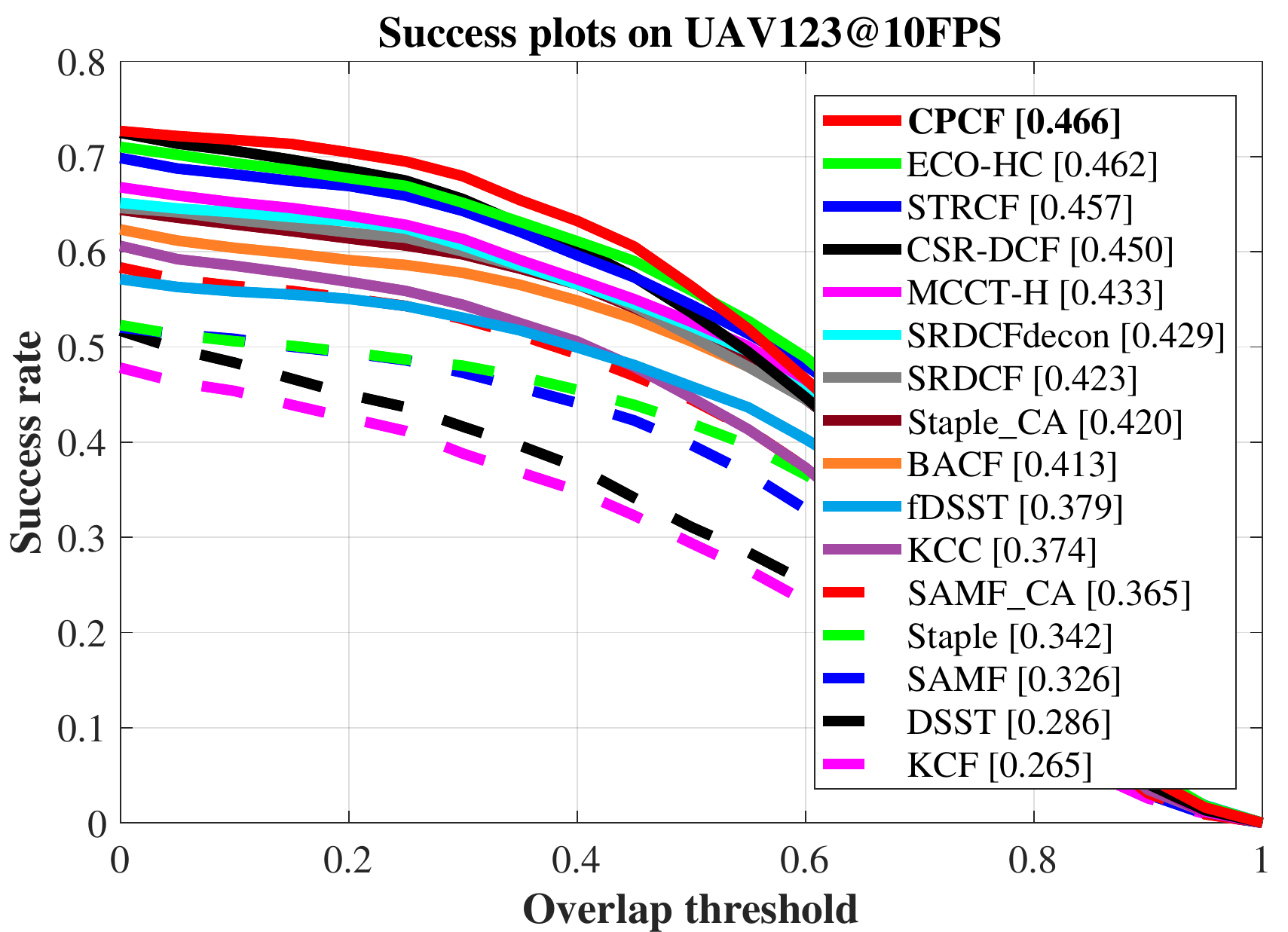}
		\end{minipage}%
		\label{fig:UAV123@10fps_performance}
	}%
	\subfigure[UAVDT benchmark]{
		\begin{minipage}[t]{0.32\textwidth}
			\includegraphics[width=1\textwidth]{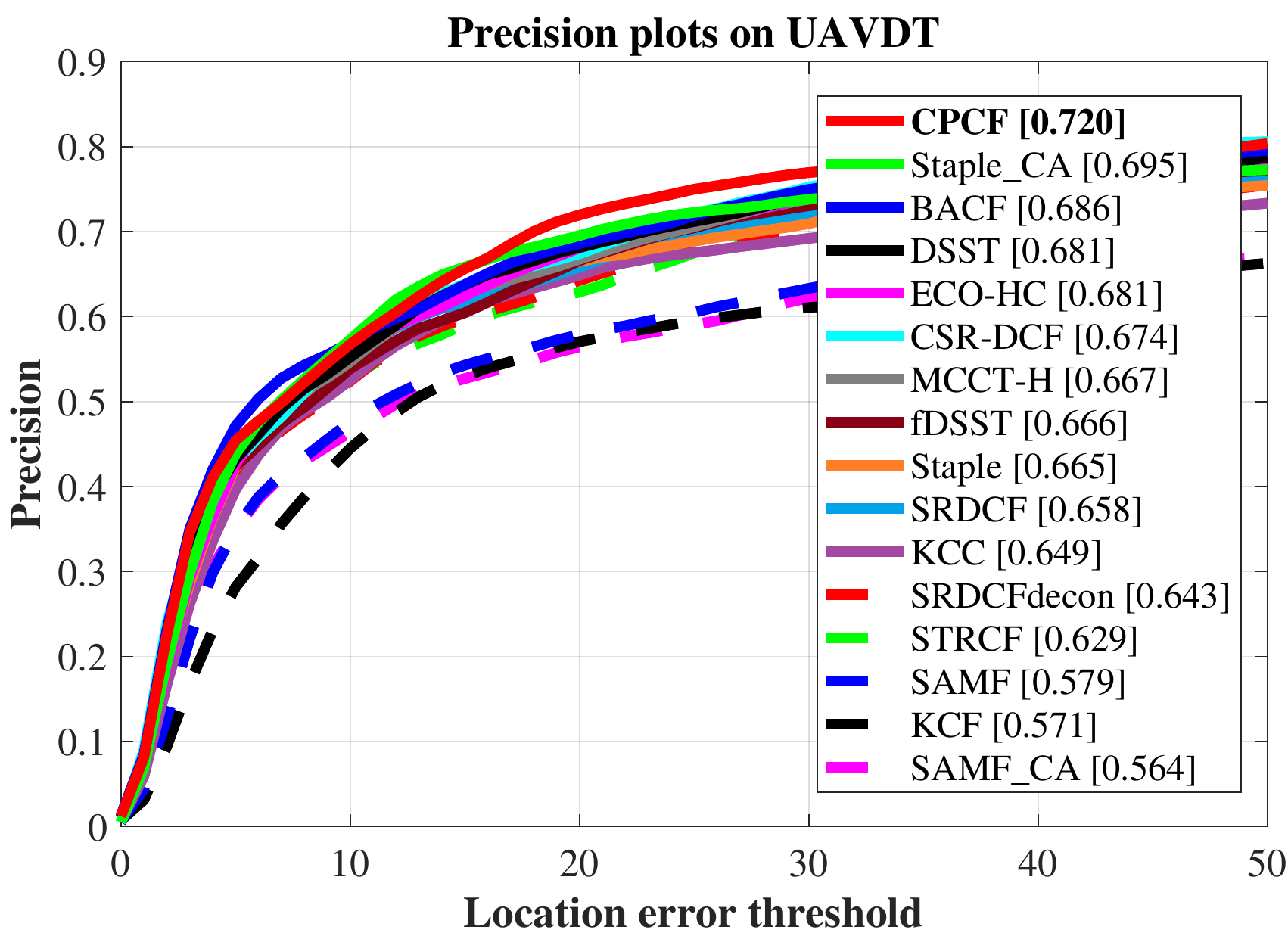}			\includegraphics[width=1\textwidth]{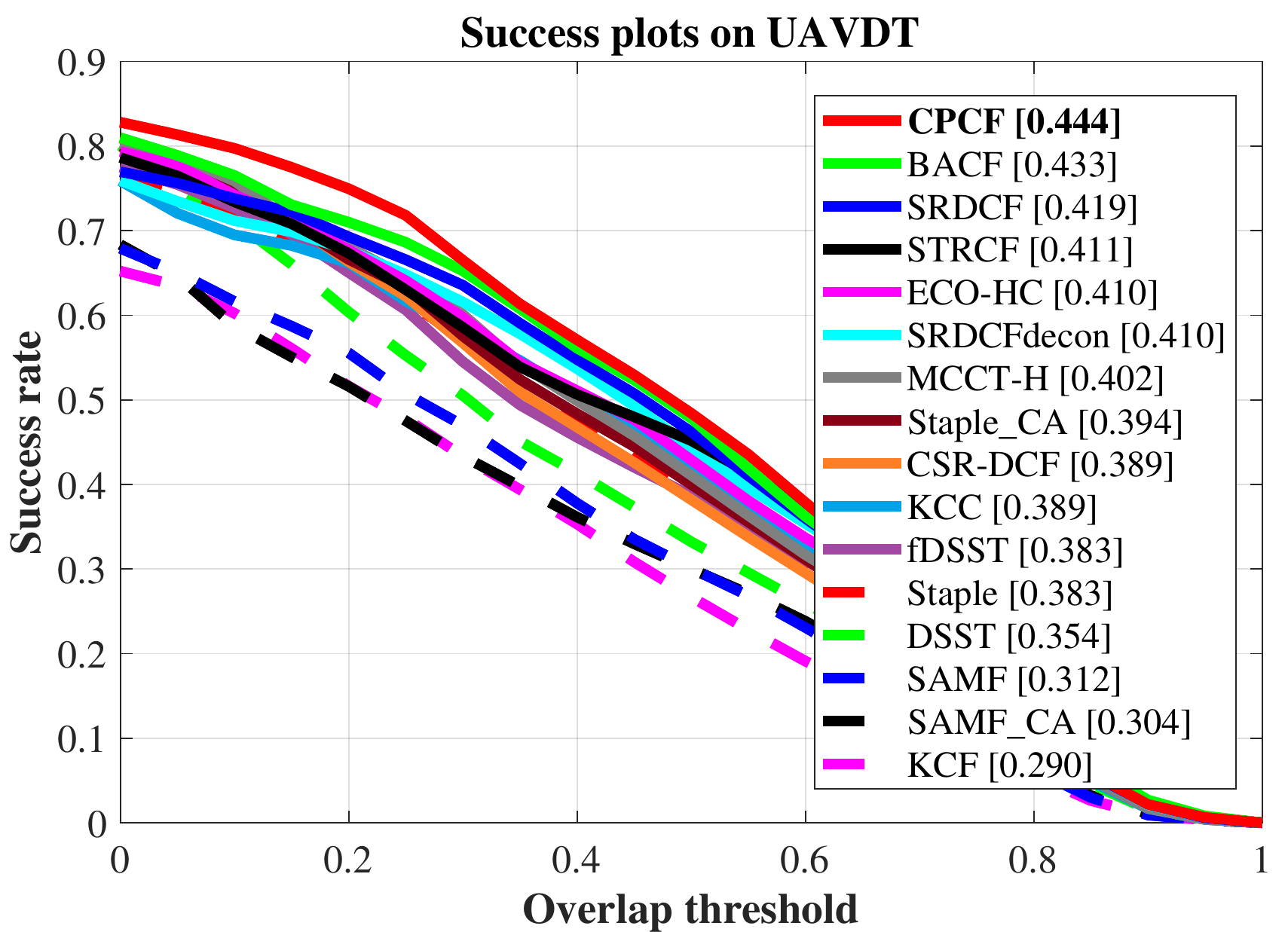}
		\end{minipage}%
		\label{fig:UAVDT_performance}
	}%
	\subfigure[DTB70 benchmark]{
		\begin{minipage}[t]{0.32\textwidth}
			\includegraphics[width=1\textwidth]{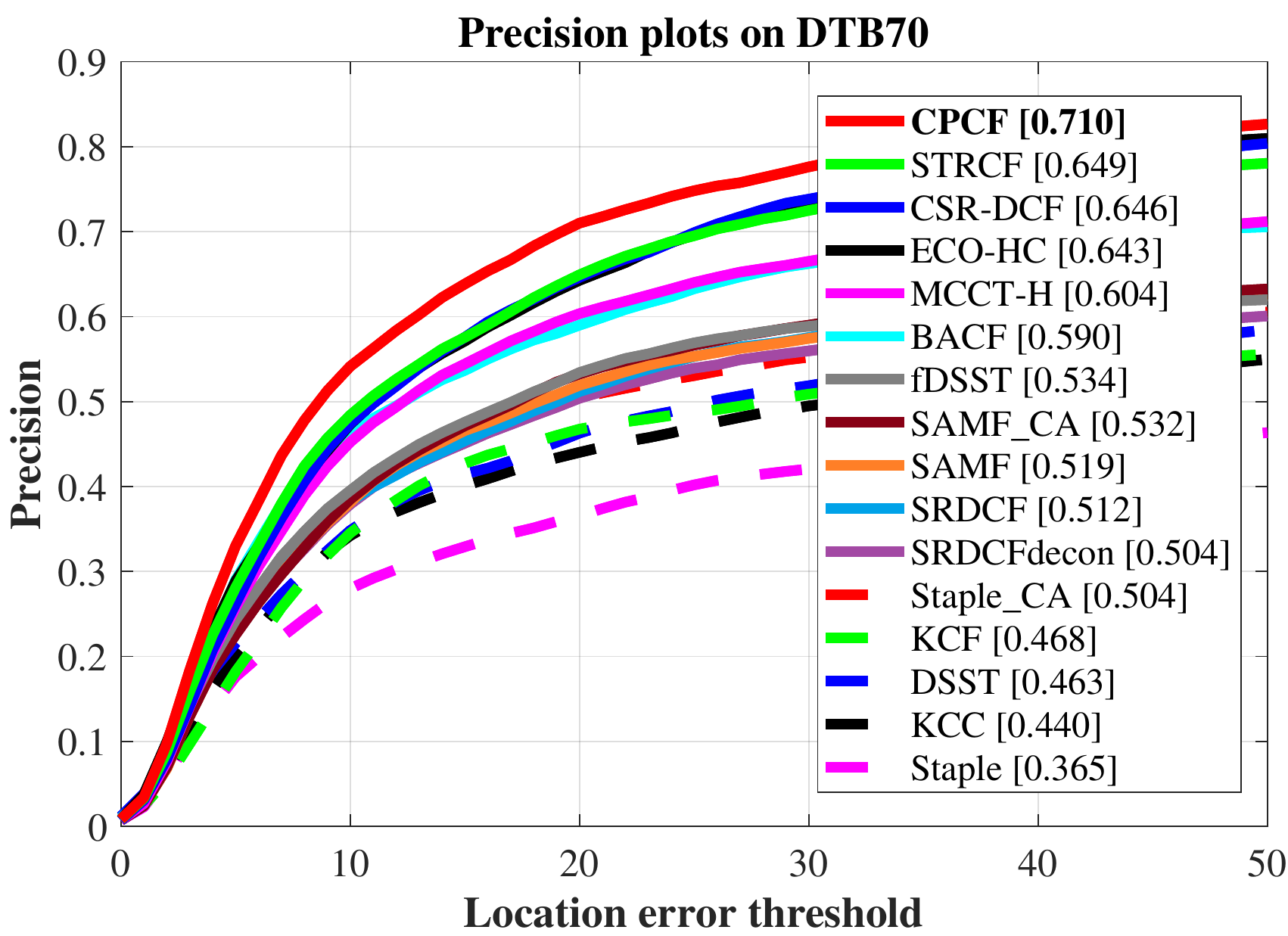}
			\includegraphics[width=1\textwidth]{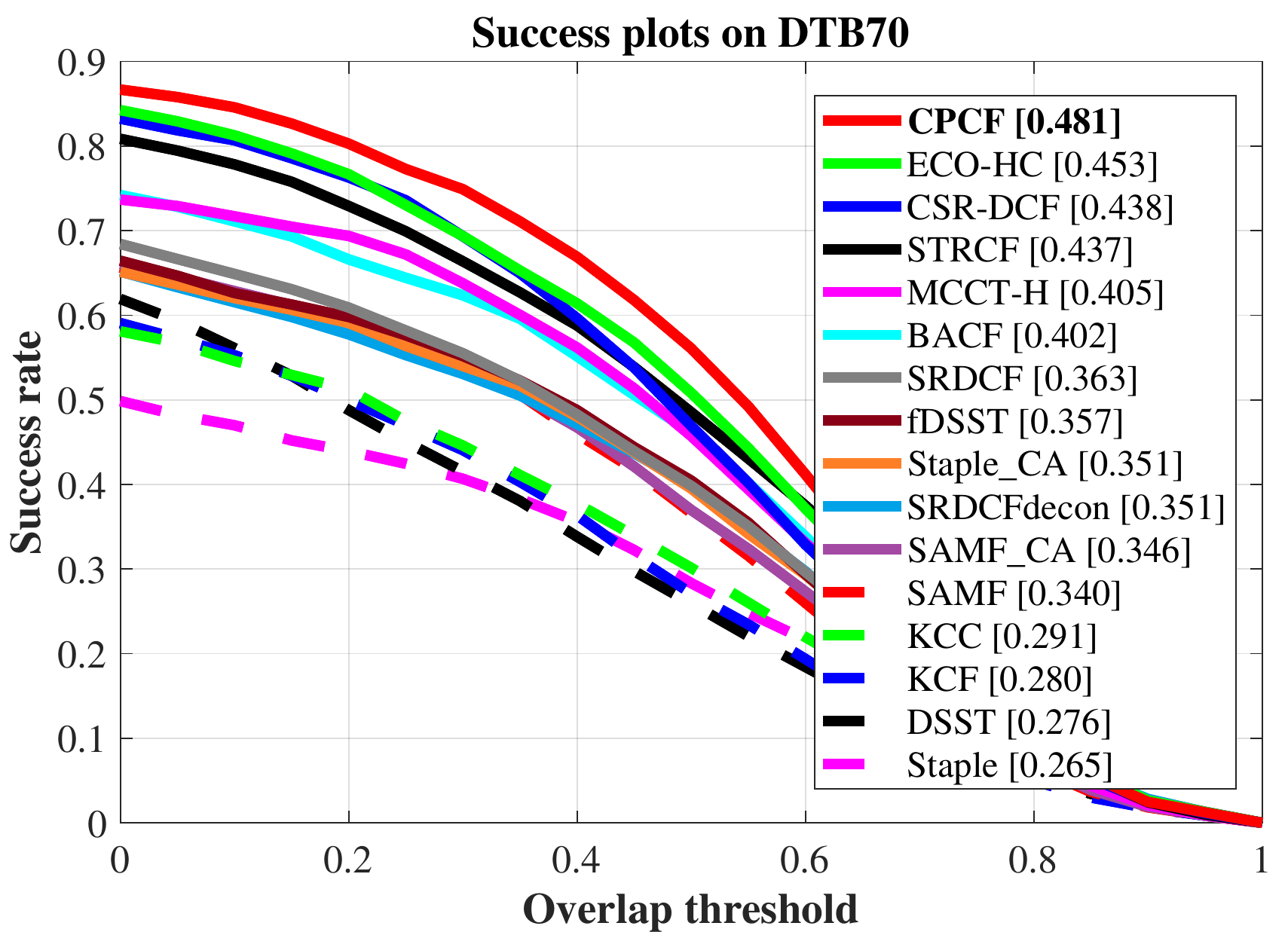}
		\end{minipage}%
		\label{fig:DT70_performance}
	}%
	\centering
	\caption{Precision and success plots of CPCF tracker, as well as all other 15 trackers based on hand-crafted features on (a) UAV123@10FPS, (b) UAVDT and (c) DTB70. 
		The proposed CPCF tracker has the best performance in terms of precision and success rate in three benchmarks.}
	\label{fig:overall_performance}
	\vspace{-5pt}
\end{figure*}

\begin{figure*}[!h]
	\centering
	\includegraphics[width=1.0\textwidth]{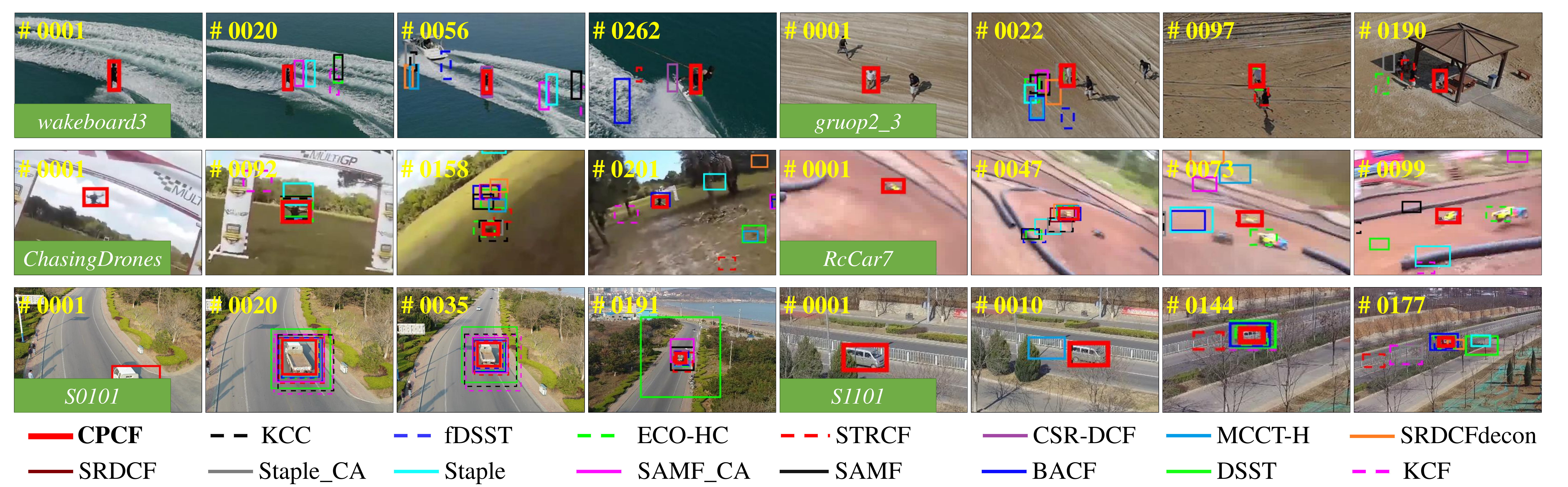} 
	\setlength{\abovecaptionskip}{-15pt} 
	\caption{Qualitative comparisons of the proposed CPCF tracker with other 15 state-of-the-art trackers. From top to bottom, the sequences are $wakeboard3$ and $group2\_3$ from UAV123@10FPS benchmark, $ChasingDrones$ and $RcCar7$ from DTB70 benchmark, $S0101$ and $S1101$ from UAVDT benchmark. Code and the UAV tracking video are: \url{https://github.com/vision4robotics/CPCF-Tracker} and \url{https://youtu.be/U0O4xI-3dl8}.}
	\label{fig:Qualitative evaluation}
	\vspace{-13pt}
\end{figure*}

\begin{figure*}[!h]\label{Attribute based comparison}
	\centering
	\setlength{\abovecaptionskip}{-1pt}
	\subfigcapskip=4pt
	\subfigure[UAV123@10FPS benchmark]{
		\begin{minipage}[t]{0.31\textwidth}
			\includegraphics[width=1\textwidth]{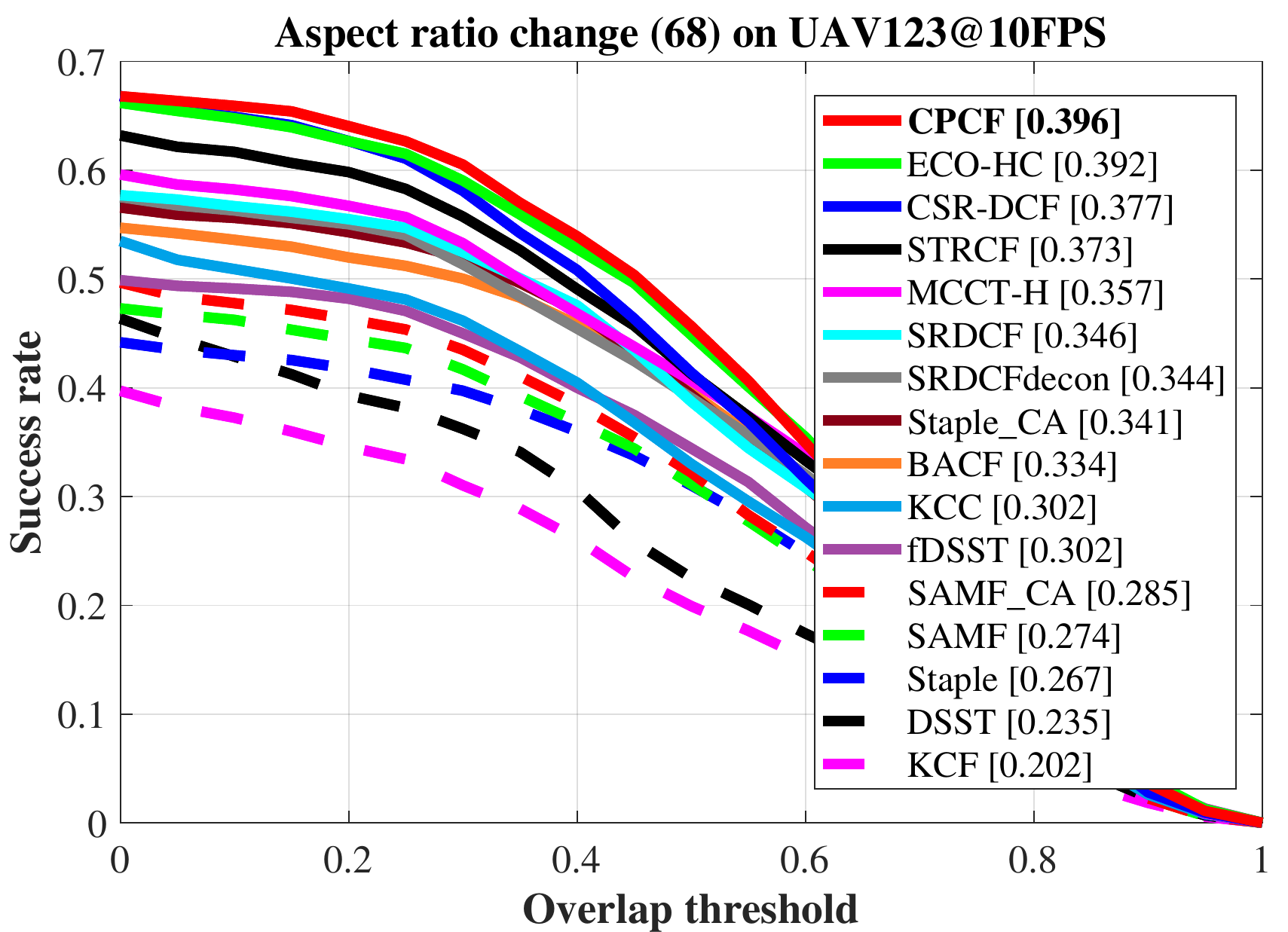}
			\includegraphics[width=1\textwidth]{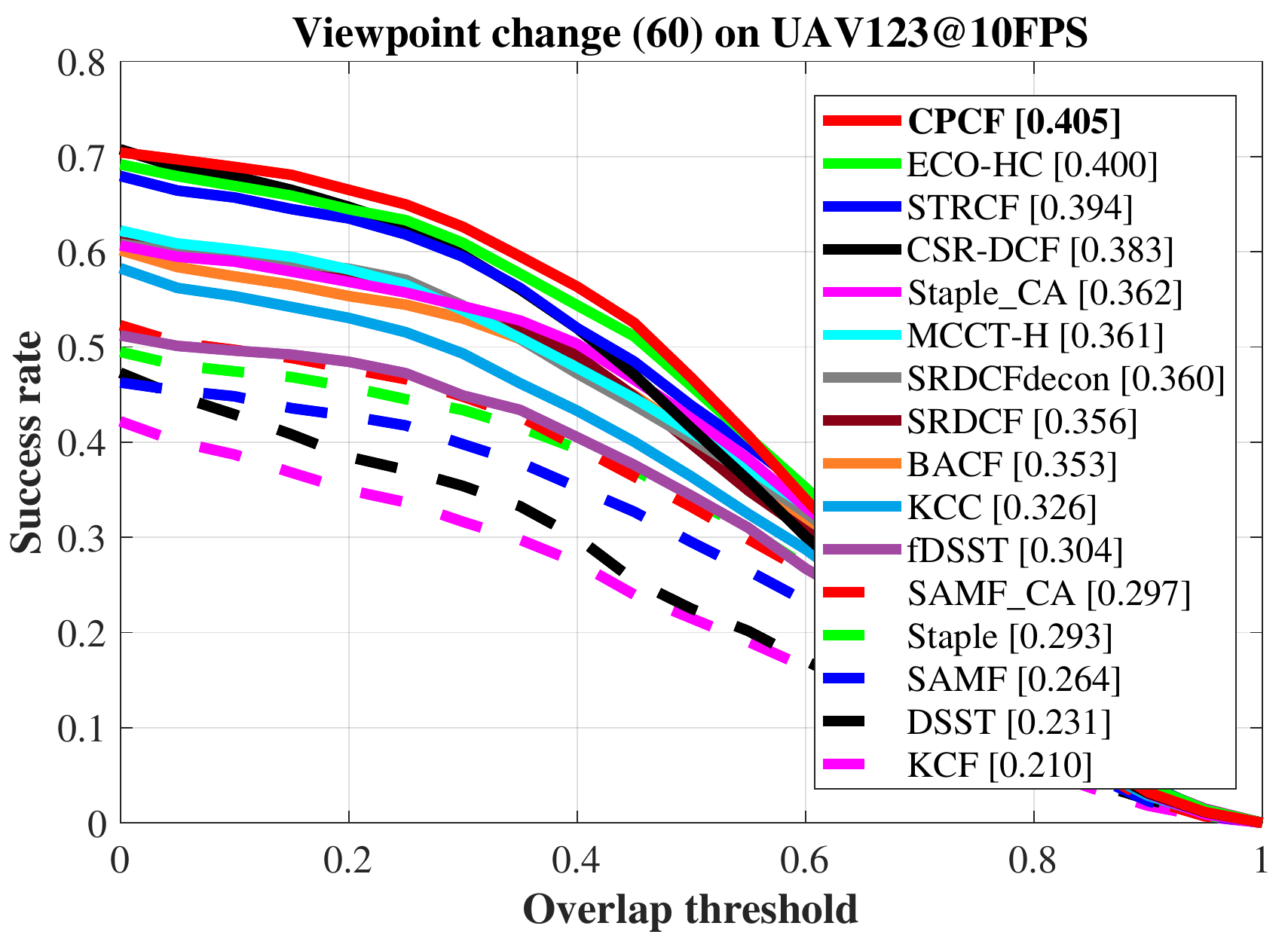}
		\end{minipage}%
		\label{fig:UAV123@10fps_attribute}
	}%
	\subfigure[UAVDT benchmark]{
		\begin{minipage}[t]{0.31\textwidth}
			\includegraphics[width=1\textwidth]{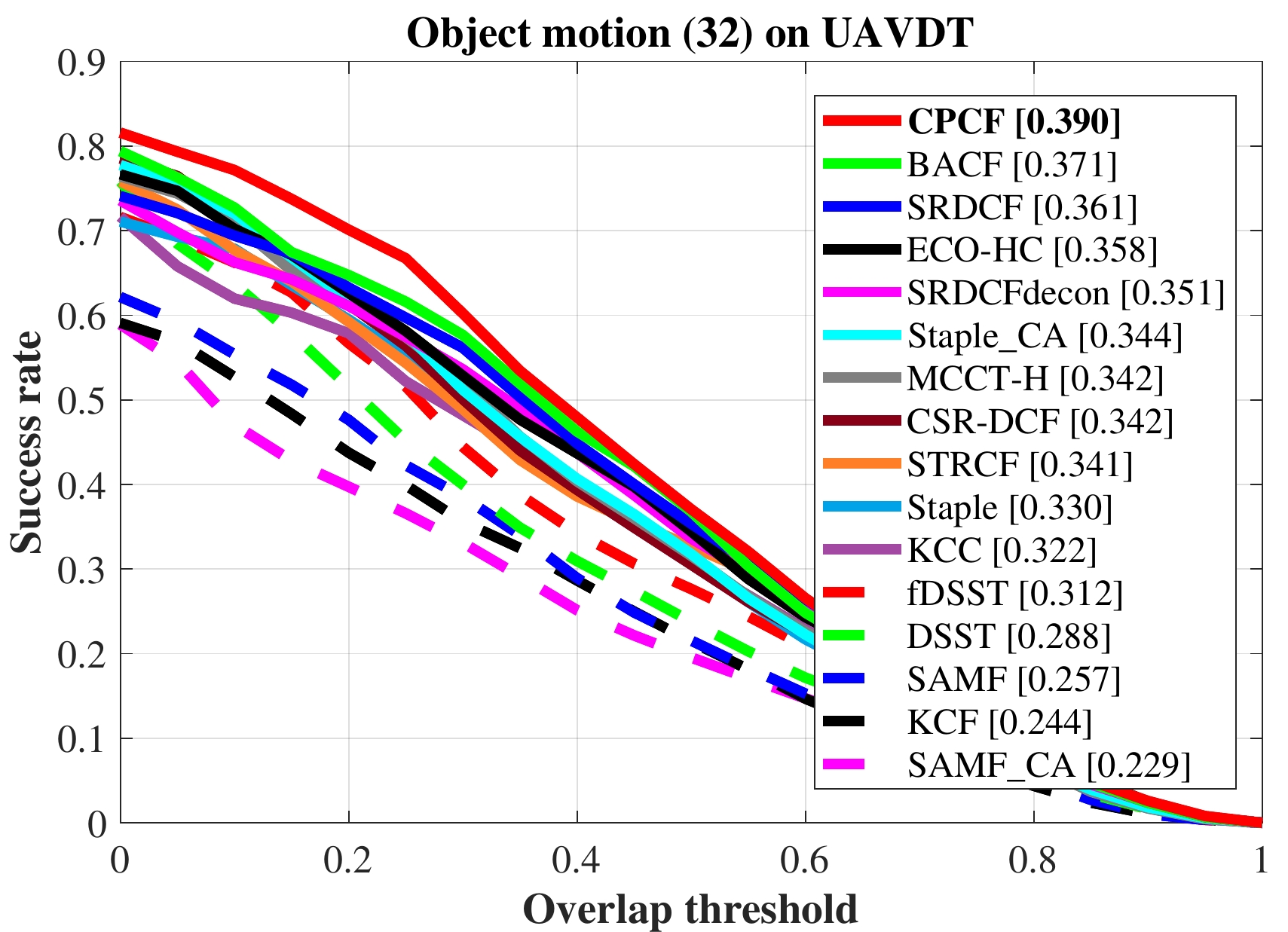}
			\includegraphics[width=1\textwidth]{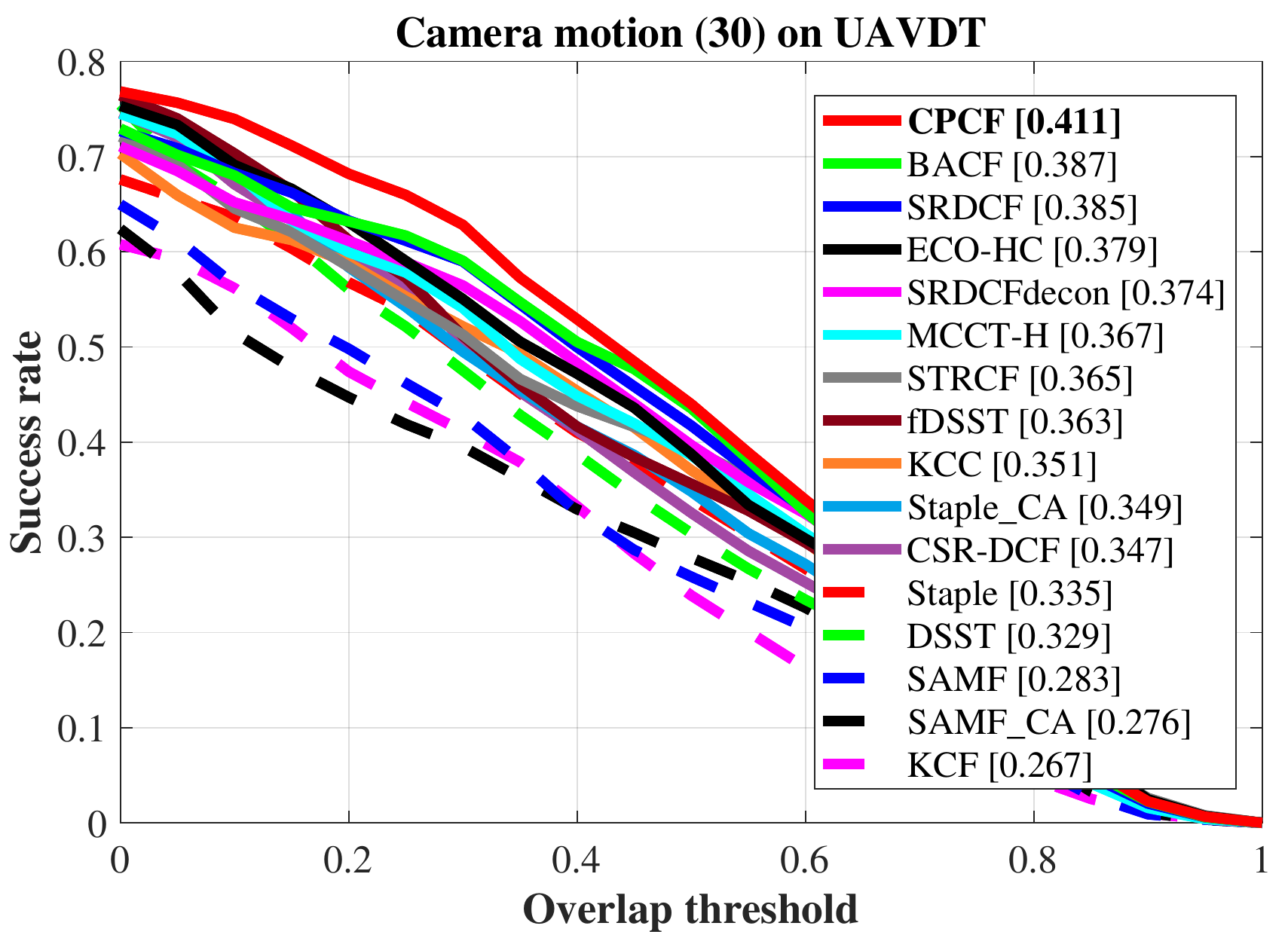}
		\end{minipage}%
		\label{fig:UAVDT_Attribute}
	}%
	\subfigure[DTB70 benchmark]{
		\begin{minipage}[t]{0.31\textwidth}
			\includegraphics[width=1\textwidth]{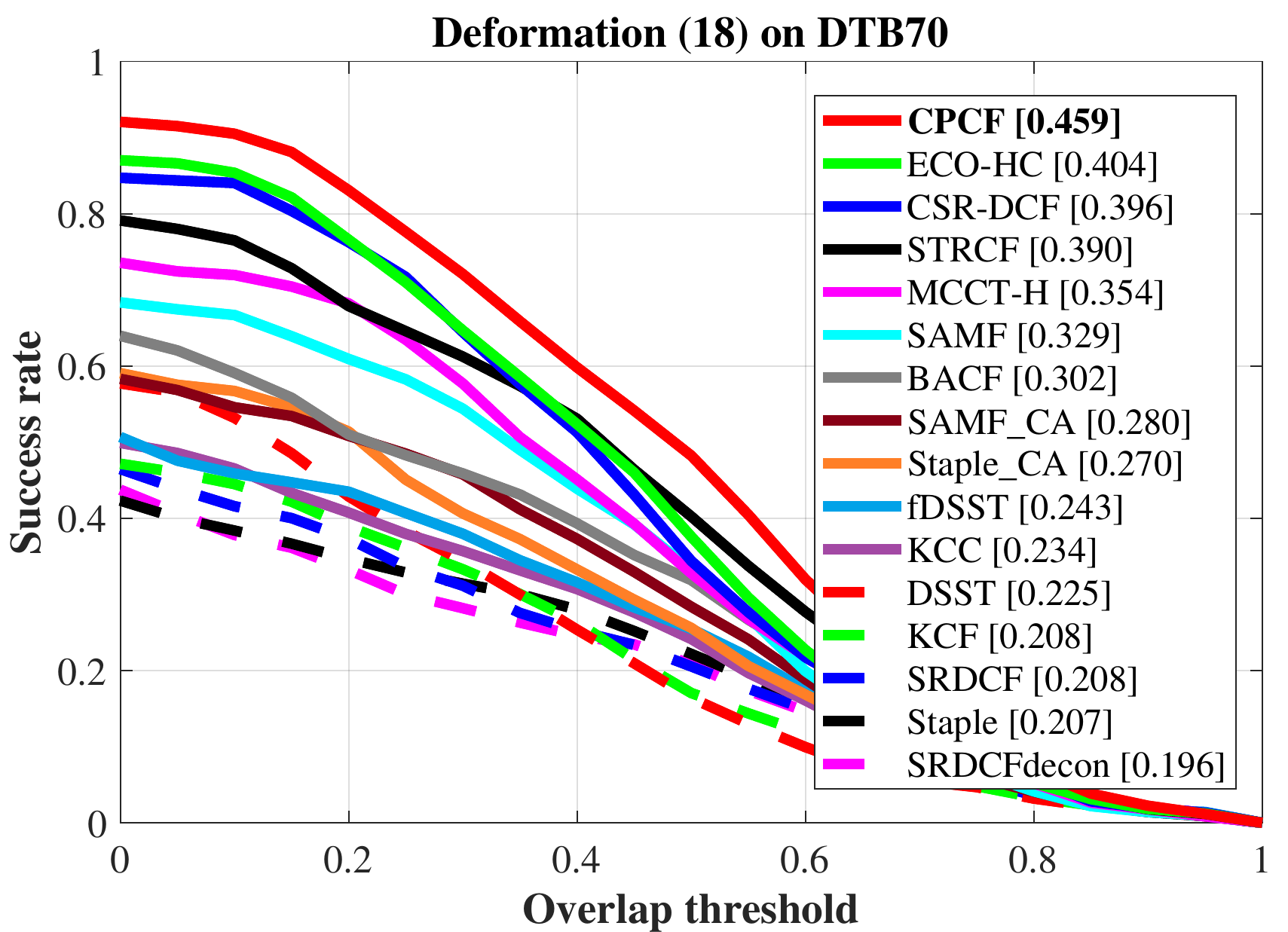}
			\includegraphics[width=1\textwidth]{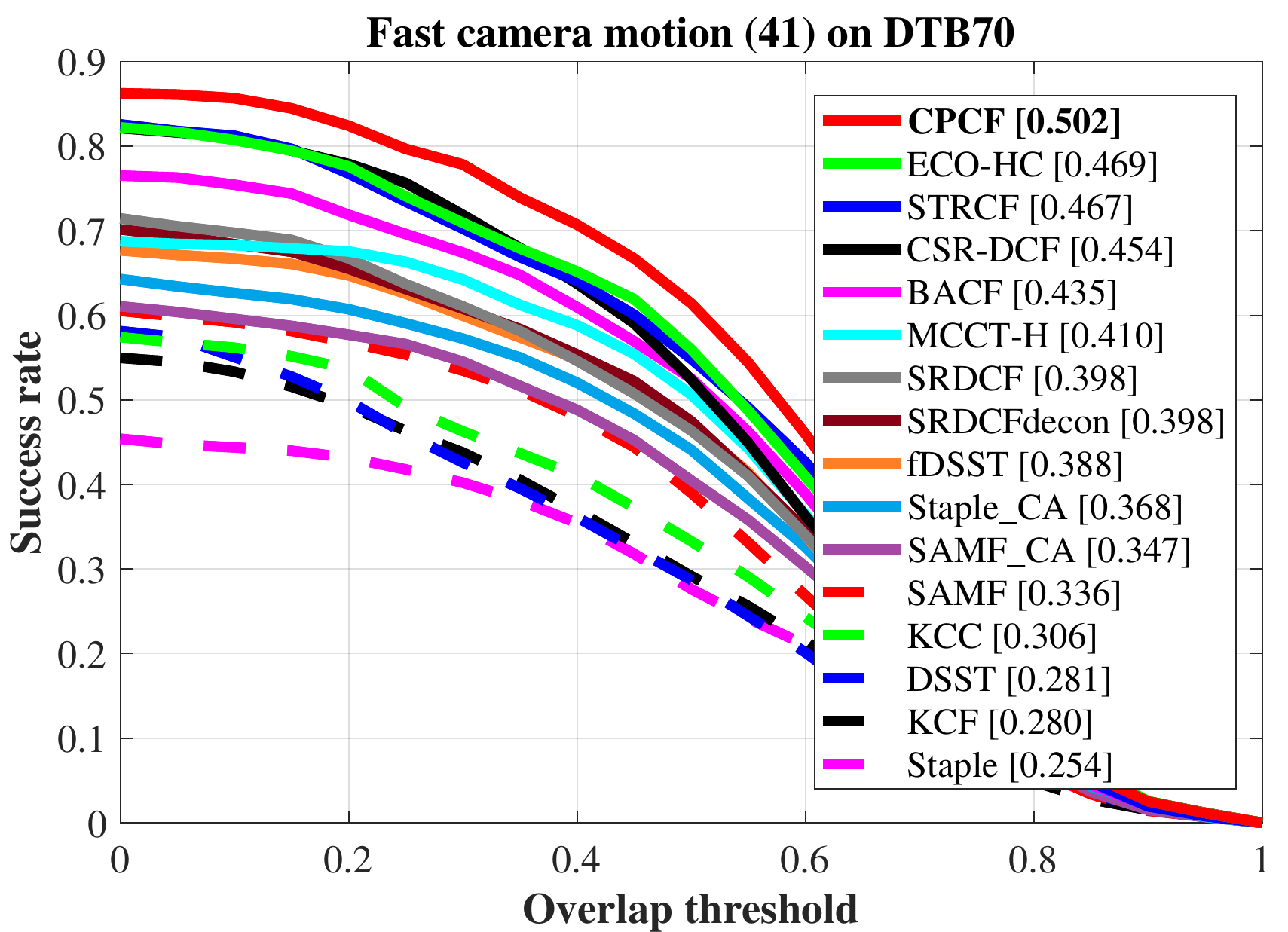}
		\end{minipage}%
		\label{fig:DT70_Attribute}
	}%
	\centering
	\caption{ Attribute-based comparisons between the CPCF tracker and other 15 trackers based on hand-crafted features on (a) UAV123@10FPS, (b) UAVDT and (c) DTB70 benchmarks. 
	All the trackers are ranked by AUC. Details of these attributes can be seen in \cite{Mueller2016ECCV,Du2018ECCV,LiAAAI2017}.
	It can be seen that the CPCF tracker has a competitive performance in these attributes across the benchmarks. 
	}
	\label{fig:Attribute comparison}
	\vspace{-10pt}
\end{figure*}

	\section{EXPERIMENTS}\label{sec:EXPERIMENT}
	\vspace{-4pt}
	In this section, the proposed CPCF tracker is evaluated comprehensively on three well-known and widely-used UAV object tracking benchmarks which are especially captured by UAV from the aerial view, \ie, UAV123@10FPS~\cite{Mueller2016ECCV}, UAVDT~\cite{Du2018ECCV}, and DTB70~\cite{LiAAAI2017}, with 243 challenging image sequences. The results are compared with 25 state-of-the-art trackers, \ie, STRCF~\cite{Li2018CVPR},  MCCT-H~\cite{Wang2018CVPR},  KCC~\cite{Wang2018AAAI},  Staple\_CA~\cite{Bertinetto2016CVPR}, SRDCF~\cite{Danelljan2015ICCV},  SAMF\_CA~\cite{Li2014ECCVws},  fDSST~\cite{Danelljan2016TPAMI},  ECO-HC~\cite{Danelljan2017CVPR},  CSR-DCF~\cite{Lukezic2017CVPR},  BACF~\cite{Galoogahi2017CVPR},  Staple~\cite{Bertinetto2016CVPR}, SRDCFdecon~\cite{Danelljan2016CVPR}, SAMF~\cite{Li2014ECCVws},  KCF~\cite{Henriques2015TPAMI},  DSST~\cite{Danelljan2014BMVC},  CFNet~\cite{Valmadre2017CVPR},  MCCT~\cite{Wang2018CVPR},  C-COT~\cite{Danelljan2016ECCV},  ECO~\cite{Danelljan2017CVPR},  IBCCF~\cite{Li2017ICCVws},  UDT+~\cite{Wang2019CVPR},  MCPF~\cite{zhang2017multi},  ADNet~\cite{yun2017action},  DeepSTRCF~\cite{Li2018CVPR}, and  TADT~\cite{Li2019CVPR}. 
	Moreover, the original evaluation criteria defined in three benchmarks respectively is adopted.

	\vspace{-5pt}
	\subsection{Implementation details}\label{subsec:impdet}
	\vspace{-4pt}
	CPCF is based on HOG~\cite{Henriques2015TPAMI} and CN~\cite{Li2014ECCVws} features. 
	The consistency penalty $\gamma$ in Eq.~(\ref{all_objective}) is set to $0.9$. 
	For the dynamic constraint strategy, ${h}_{min}$ and ${h}_{max}$ in Eq.~(\ref{Eq:mag&mcer}) are set to $0.6$ and $1.2$, respectively.
	The normalized coefficient $\alpha$ and the weight coefficient $\beta$ in Eq.~(\ref{Eq:mag&mcer}) and Eq.~(\ref{PSRM}) are set to 50 and 100, respectively. The learning rate $\eta$ in Eq.~(\ref{modelupdate}) is set to $0.042$. 
	All the 26 trackers are performed with MATLAB R2018a on a computer with an i7-8700K CPU (3.7GHz), 32GB RAM and Nvidia GeForce RTX 2080. 
	Note that the CPCF tracker is tested on a single CPU.

	\vspace{-0.5em}
	\subsection{Comparison with hand-crafted based trackers}
	\label{subsec:overall}
	\vspace{-4pt}
	\textbf{Quantitative evaluation:} 
	As shown in Fig.~\ref{fig:stplot}, the average overall performance of CPCF and other 15 state-of-the-art trackers utilizing hand-crafted features is demonstrated.
	The CPCF tracker has surpassed all compared trackers on three UAV benchmarks.
	In addition, the overall performance of each benchmark is shown in Fig.~\ref{fig:overall_performance}. 
	Concretely, on UAV123@10FPS benchmark, CPCF (0.661) outperforms the second-best CSR-DCF (0.643) and the third-best ECO-HC (0.634) by 1.8\% and 2.7\%, respectively in precision, and has an advantage of 0.4\% and 0.9\% over the second-best (ECO-HC, 0.462) and the third-best  (STRCF, 0.457), respectively in AUC. 
	On UAVDT benchmark, CPCF (0.720, 0.444) surpasses the second-best (Staple\_CA, 0.695) and the third-best (BACF, 0.686) by 2.5\% and 3.4\%, respectively in precision, as well as an advancement of 1.1\% and 2.5\% over BACF (0.433) and SRDCF (0.419), respectively in AUC. 
	On DTB70 benchmark, CPCF (0.710, 0.481) is followed by STRCF (0.649) and CSR-DCF (0.646) in precision, and by ECO-HC (0.453) and CSR-DCF (0.438) in AUC. 
	\indent\textbf{Qualitative evaluation:} 
	The comparisons of our approach with other trackers are visualized in Fig. \ref{fig:Qualitative evaluation}.
	It can be seen that the CPCF tracker performs satisfactorily in different challenging scenarios.\\
	\indent\textbf{Speed comparison:} 
	The speed of CPCF is sufficient for real-time UAV tracking applications, and the comparison can be seen in Tabel~\ref{tab:hand speed}.
	\\
	\indent\textbf{Attribute based comparison:} 
	To better demonstrate the ability of CPCF to respond to different challenges, the three benchmarks classify them into different attributes, such as viewpoint change and object motion. 
	Examples of attribute-based comparisons are shown in Fig. \ref{fig:Attribute comparison}, which are ranked by AUC.
	The comparisons demonstrate that the CPCF tracker has a great improvement over its baseline SRDCF, due to the pursuit of consistency across frames.
	\begin{table}[htbp]
		\centering
		\caption{The average frame per second (FPS) comparison between the CPCF tracker and other 15 trackers on all three benchmarks \cite{Du2018ECCV,Mueller2016ECCV,LiAAAI2017}. \textbf{\textcolor[rgb]{1,0,0}{Red}, \textcolor[rgb]{0,1,0}{green}} and \textbf{\textcolor[rgb]{0,0,1}{blue}} color denote the first, second and third place respectively. Note that the presented CPCF tracker is tested on a single CPU.}
		\vspace{-8pt}
		\setlength{\tabcolsep}{0.5mm}{
			\begin{tabular}{ccc|ccc}
				\toprule
				\toprule
				Tracker & FPS   & Venue & Tracker & FPS   & Venue \\
				\midrule
				\textbf{CPCF}  & \textbf{42.95}  & \textbf{Ours}  & CSR-DCF & 12.09  & CVPR'17 \\
				MCCT-H & 59.72  & CVPR'18 & BACF  & 56.04  & ICCV'17 \\
				KCC   & 46.12  & AAAI'18 & Staple & 65.40  & CVPR'16 \\
				Staple\_CA & 58.86  & CVPR'17 & SRDCFdecon & 7.48  & CVPR'16 \\
				SRDCF & 14.01  & ICCV'17 & SAMF  & 12.76  & ECCV'14 \\
				SAMF\_CA & 11.66  & CVPR'17 & KCF   & \textcolor[rgb]{ 1,  0,  0}{651.06 } & TPAMI'14 \\
				fDSST & \textcolor[rgb]{ 0,  1,  0}{168.06 } & TPAMI'17 & DSST  & \textcolor[rgb]{ 0,  0,  1}{106.49 } & BMVC'14 \\
				ECO-HC & 69.33  & CVPR'17 & STRCF & 28.51  & CVPR'18 \\
				\bottomrule
				\bottomrule
			\end{tabular}%
		}%
		\label{tab:hand speed}%
		\vspace{4pt}
	\end{table}%
	\begin{table*}[htbp]
		\centering
		\caption{The frame per second (FPS), precision and AUC comparison between CPCF and deep-based trackers on UAVDT \cite{Du2018ECCV} benchmark. \textbf{\textcolor[rgb]{1,0,0}{Red}, \textcolor[rgb]{0,1,0}{green}} and \textbf{\textcolor[rgb]{0,0,1}{blue}} color denote the first, second and third place respectively. 
			The presented CPCF tracker performs favorably compared with other deep-based trackers.  }
		\vspace{-8pt}
		\setlength{\tabcolsep}{1.2mm}{\begin{tabular}{cccccccccccc}
				\toprule
				\toprule
				Tracker & \textbf{CPCF} & CFNet & MCCT  & C-COT & ECO   & TADT  & IBCCF & UDT+  & MCPF  & ADNet & DeepSTRCF \\
				\midrule
				FPS   & \textcolor[rgb]{ 0,  1,  0}{48.29}  & \textcolor[rgb]{ 0,  0,  1}{41.05 } & 8.60  & 1.10  & 16.38  & {32.48 } & 3.39  & \textcolor[rgb]{ 1,  0,  0}{60.42 } & 3.63  & 7.55  & 6.61  \\
				\midrule
				Precison & \textcolor[rgb]{ 1,  0,  0}{0.720 } & 0.680  & 0.671  & 0.656  & \textcolor[rgb]{ 0,  1,  0}{0.700 } & 0.677  & 0.603  & \textcolor[rgb]{ 0,  0,  1}{0.697 } & 0.660  & 0.683  & 0.667  \\
				\midrule
				AUC   & \textcolor[rgb]{ 0,  1,  0}{0.444} & 0.428 & 0.437 & 0.406 & \textcolor[rgb]{ 1,  0,  0}{0.454} & 0.431 & 0.388 & 0.416 & 0.399 & 0.429 & \textcolor[rgb]{ 0,  0,  1}{0.437} \\
				\midrule
				Venue & Ours  & 2017CVPR & 2018CVPR & 2016ECCV & 2017CVPR & 2019CVPR & 2017CVPR & 2019CVPR & 2017CVPR & 2017CVPR & 2018CVPR \\
				\midrule
				GPU   & $ \text{\sffamily X} $     & $\checkmark$     & $\checkmark$     & $\checkmark$     & $\checkmark$     & $\checkmark$     & $\checkmark$     & $\checkmark$     & $\checkmark$     & $\checkmark$     & $\checkmark$ \\
				\bottomrule
				\bottomrule
				
			\end{tabular}%
		}

		\label{tab:fpslabel}%
		\vspace{-16pt}
	\end{table*}%
	\vspace{-2pt}
	\subsection{Comparison with deep-based trackers}
	\label{subsec:comp deep trackers}
	\vspace{-5pt}
	In order to fully reflect the performance of CPCF tracker for UAV tracking applications, the CPCF tracker is also compared with deep-based trackers on the UAVDT benchmark. 
	In terms of precision, success rate and speed, the CPCF tracker has performed favorably against other state-of-the-art compared trackers. 
	The comparisons are shown in Table \ref{tab:fpslabel}. Note that all deep-based trackers are tested on the GPU, while the CPCF tracker is tested on a single CPU.
	
	\vspace{-7pt}
	\section{CONCLUSIONS}\label{sec:CONCLUSIONS}
	\vspace{-6pt}
	In this work, a novel approach with dynamic consistency pursued correlation filters, \ie, the CPCF tracker, is proposed. 
	Generally, by exploiting the consistency across frames, rich temporal information in the response maps is introduced to enhance the discriminative power of the tracker. 
	Besides, a dynamic consistency constraint is proposed to strengthen the adaptability in complex situations.
	Considerable experiments are conducted on three UAV object tracking benchmarks. The experimental results verify the outstanding performance of the presented tracker compared with 25 state-of-the-art trackers. 
	Moreover, the CPCF tracker obtains a real-time speed ($\sim$43FPS) on a single CPU.
	From our view, the temporal information behind response maps is further explored by consistency representation, which can contribute to the object tracking on-board UAV. 
	
	\vspace{-5pt}
	\section*{APPENDIX}
	\label{APPENDIX}
		\vspace{-3pt}
	Solving process from (\ref{gkn}) to (\ref{equ:subprobg}) :
	\vspace{-5pt}
	\begin{small}
		\begin{equation}
		\begin{split} 
		\small
		&\hat { \mathbf { g } }_{k+1}^* ( n ) ^ { } =  \mathop{\arg \min} _ { \hat { \mathbf { g } }_{k} ( n ) } \{ \frac { 1 } { 2 } \Vert \hat { \mathbf { y } } ( n ) - \hat { \mathbf { x }}_k  ( n ) ^ { \top } \hat { \mathbf { g } }_k ( n ) \Vert_2^2  \\ 
		&\quad\quad\quad\quad\quad\quad\quad+ \frac { \gamma } { 2 } \Vert \hat { \mathbf { l } }_k ( n ) - \hat { \mathbf { x }}_k^r ( n ) ^ { \top } \hat { \mathbf { g } }_k ( n ) \Vert_2^2  \\ 
		&\quad\quad\quad\quad\quad \quad\quad+\hat { \zeta } ( n ) ^ { \top } ( \hat { \mathbf { g } }_k ( n ) - \hat { \mathbf { w } }_k ( n ) ) \\ 
		&\quad\quad\quad\quad\quad\quad\quad + \frac { \nu } { 2 } \| \hat { \mathbf { g } }_k ( n ) - \hat { \mathbf { w } }_k ( n ) \| _ { 2 } ^ { 2 } \} 
		\end{split}
		\ ,
		\end{equation}
	\end{small}
	\vspace{-5pt}
	\begin{small}
		\begin{equation}
		\begin{aligned}
		\text{Let} \  G & =  \frac { 1 } { 2 } \Vert \hat { \mathbf { y } } ( n ) - \hat { \mathbf { x }}_k  ( n ) ^ { \top } \hat { \mathbf { g } }_k ( n ) \Vert_2^2  \\ 
		&+ \frac { \gamma } { 2 } \Vert \hat { \mathbf { l } }_k ( n ) - \hat { \mathbf { x }}_k^m ( n ) ^ { \top } \hat { \mathbf { g } }_k ( n ) \Vert_2^2  \\ 
		& +\hat { \zeta } ( n ) ^ { \top } ( \hat { \mathbf { g } }_k ( n ) - \hat { \mathbf { w } }_k ( n ) ) \\ 
		& + \frac { \nu } { 2 } \| \hat { \mathbf { g } }_k ( n ) - \hat { \mathbf { w } }_k ( n ) \| _ { 2 } ^ { 2 } 
		\end{aligned}
		\  ,
		\end{equation}
	\end{small}
	\vspace{-6pt}
\begin{small}
	\begin{equation}
	\begin{aligned} 
	\frac { \partial \mathrm { G }^ { \top } } { \partial  { \hat{\mathbf { g }}_k }  ( n ) } = & - \left( \hat { \mathbf { y } } ( n ) - \hat { \mathbf { x } } _k {  } ( n ) ^ { \top } \hat { \mathbf { g }_k }  ( n ) \right) ^ { \top } \hat { \mathbf { x } } _ { k } ( t ) ^ { \top } \\ &  - \gamma\left( \hat { \mathbf { l } }_k ( n ) - \hat { \mathbf { x } }_k^m {  } ( n ) ^ { \top } \hat { \mathbf { g } }_k  ( n ) \right) ^ { \top } \hat { \mathbf { x } } _ { k }^m ( t ) ^ { \top } \\ & + \hat { \zeta }  ( n ) ^ { \top } + \nu  \left( \hat { \mathbf { g } } ( n ) - \hat { \mathbf { w } }  ( n ) \right) ^ { \top } =0 
	\end{aligned}
	\ ,
	\end{equation}
\end{small}
\vspace{-1pt}
	then Eq. (\ref{equ:subprobg}) is obtained. \hfill $\blacksquare$
	\vspace{-8pt}
	\section*{ACKNOWLEDGMENT}
	\vspace{-5pt}
	This work is supported by the National Natural Science Foundation of China (No. 61806148).
	
	
	{	\vspace{-2pt}
		\small
			\vspace{-5pt}
		\bibliographystyle{IEEEtran}
		\bibliography{van}
	}
	
\end{document}